\documentclass[sigconf]{acmart}
\usepackage{balance}
\usepackage{flushend}
\usepackage{graphicx}
\usepackage{subcaption}
\usepackage{multicol}
\usepackage{multirow}
\usepackage{colortbl}
\AtBeginDocument{%
  }

\copyrightyear{2025}
\acmYear{2025}
\setcopyright{cc}
\setcctype{by}
\acmConference[CIKM '25]{Proceedings of the 34th ACM International Conference on Information and Knowledge Management}{November 10--14, 2025}{Seoul, Republic of Korea}
\acmBooktitle{Proceedings of the 34th ACM International Conference on Information and Knowledge Management (CIKM '25), November 10--14, 2025, Seoul, Republic of Korea}\acmDOI{10.1145/3746252.3761379}
\acmISBN{979-8-4007-2040-6/2025/11}

\begin{document}

\title{Efficient Knowledge Graph Unlearning with \\Zeroth-order Information}

\author{Yang Xiao}
\email{yax3417@utulsa.edi}
\affiliation{%
  \institution{The University of Tulsa}
  \city{Tulsa}
  \state{Oklahoma}
  \country{USA}
}

\author{Ruimeng Ye}
\email{ruy9945@utulsa.edu}
\affiliation{%
  \institution{The University of Tulsa}
  \city{Tulsa}
  \state{Oklahoma}
  \country{USA}
}

\author{Bohan Liu}
\email{bohanli2@andrew.cmu.edu}
\affiliation{%
  \institution{Carnegie Mellon University}
  \city{Pittsburgh}
  \state{Pennsylvania}
  \country{USA}}

\author{Xiaolong Ma}
\email{xiaolongma@arizona.edu}
\affiliation{%
  \institution{The University of Arizona}
  \city{Tucson}
  \state{Arizona}
  \country{USA}
}

\author{Bo Hui}
\email{bo-hui@utulsa.edu}
\affiliation{%
  \institution{The University of Tulsa}
  \city{Tulsa}
  \state{Oklahoma}
  \country{USA}
}


\renewcommand{\shortauthors}{Yang Xiao et al.}

\begin{abstract}
% Due to regulatory guidelines such as the Right to be Forgotten, there is an increasing demand for the capability to proactively remove training data and eliminate its influence from a trained model. While fully retraining the model is computationally infeasible, various machine unlearning algorithms have been proposed to approximate the influence of training data removed without retraining. 
Due to regulations like the Right to be Forgotten, there is growing demand for removing training data and its influence from models. Since full retraining is costly, various machine unlearning methods have been proposed.
In this paper, we firstly present an efficient knowledge graph (KG) unlearning algorithm. We remark that KG unlearning is nontrivial due to the distinctive structure of KG and the semantic relations between entities. Also, unlearning by estimating the influence of removed components incurs significant computational overhead when applied to large-scale knowledge graphs. To this end, we define an influence function for KG unlearning and propose to approximate the model's sensitivity without expensive computation of first-order and second-order derivatives for parameter updates. Specifically, we use Taylor expansion to estimate the parameter changes caused by data removal. Given that the first-order gradients and second-order derivatives dominate the computational load, we use the Fisher matrices and zeroth-order optimization to approximate the inverse-Hessian
vector product without constructing the computational
graphs. Our experimental
results demonstrate that the proposed method outperforms other state-of-the-art graph unlearning baselines significantly in terms of unlearning efficiency and unlearning quality. Our code is
released at~\url{https://github.com/NKUShaw/ZOWFKGIF}. 
\end{abstract}

\keywords{Knowledge Graph; Influence Function; Zeroth Optimization}
\begin{CCSXML}
<ccs2012>
<concept>
<concept_id>10002978.10003018</concept_id>
<concept_desc>Security and privacy~Database and storage security</concept_desc>
<concept_significance>500</concept_significance>
</concept>
</ccs2012>
\end{CCSXML}

\ccsdesc[500]{Security and privacy~Database and storage security}
\maketitle

\section{Introduction}

\begin{figure}[t]\label{image:Unlearning}
    \centering
    \includegraphics[width=0.97\linewidth]{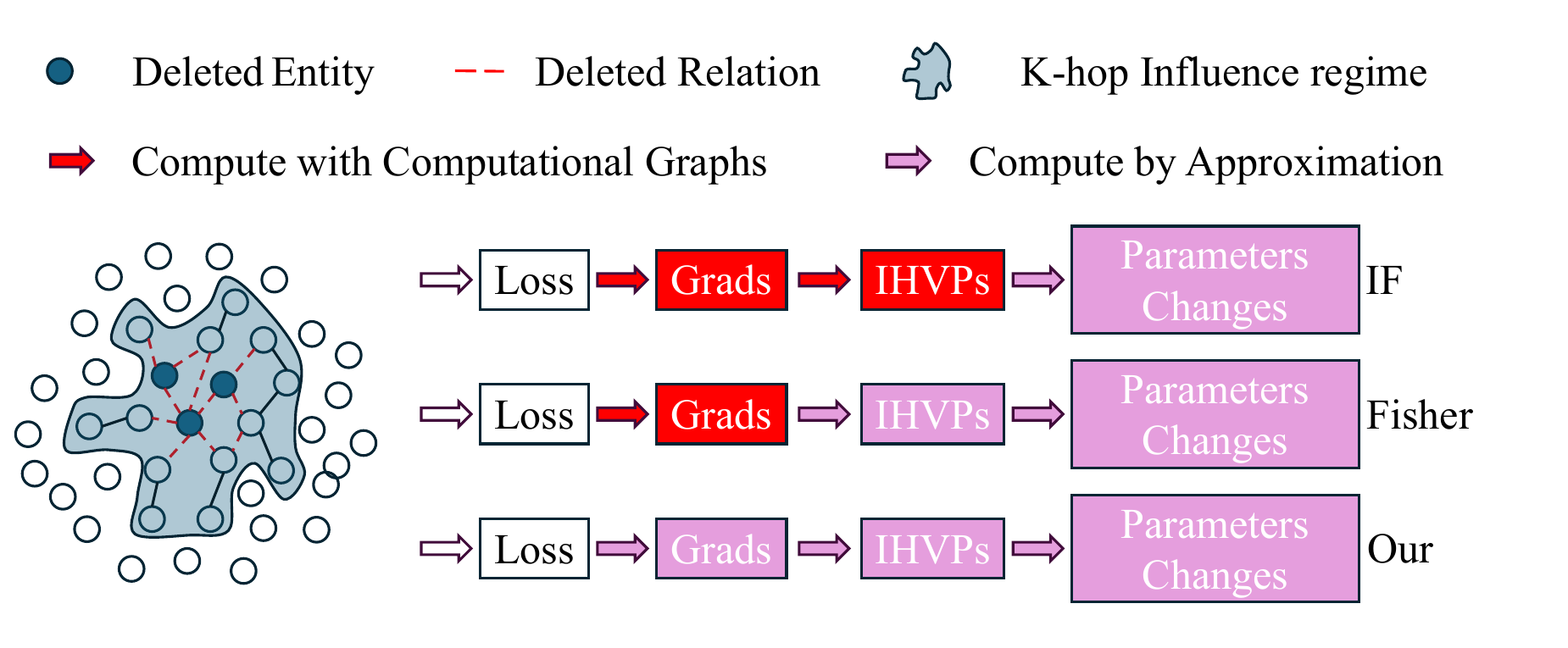}
    \caption{Influence Function Approximation}
    \label{fig: workflow}
\end{figure}

% With massive data being used in the machine learning (ML) model training process, data privacy and security have gained significant attention. Laws such as the European Union's General Data Protection Regulation (GDPR), the California Consumer Privacy Act (CCPA), and Canada's proposed Consumer Privacy Protection Act (CPPA) oversee the use of personal data in machine learning and grant users the \textit{Right to be Forgotten}~\cite{biega2021reviving, regulation2016regulation, oag2021ccpa}.
With massive data being used in the machine learning (ML) model training process, data privacy and security have gained significant attention. Laws such as the European Union's General Data Protection Regulation (GDPR), the California Consumer Privacy Act (CCPA), and Canada's proposed Consumer Privacy Protection Act (CPPA) oversee the use of personal data in machine learning and grant users the \textit{Right to be Forgotten}~\cite{biega2021reviving, regulation2016regulation, oag2021ccpa}. This right entitles individuals to request the removal of their personal data from databases, including those used for training ML models. However, traditional ML pipelines are not designed to easily accommodate such deletion requests, as once data is integrated into a model, its influence is deeply embedded in learned parameters through complex and non-linear transformations.

This mismatch between legal requirements and technical capabilities has sparked growing interest in developing Machine Unlearning~\cite{cao2015towards}, the techniques that aims to remove the influence of deleted data from trained models upon certified requests without retraining from scratch.
While retraining a new model on remaining data is intuitive~\cite{ginart2019making, said2023survey}, it is computationally impractical for frequent deletions, especially with large-scale models. To address this, methods for exact or approximate unlearning have emerged~\cite{DBLP:conf/aaai/GravesNG21,goel2022towards, DBLP:conf/nips/KurmanjiTHT23,DBLP:conf/aaai/MoonC024, DBLP:conf/aaai/ChaCHLML24}, targeting specific data types like face images~\cite{DBLP:conf/cvpr/ChenGL0W23}, natural language~\cite{DBLP:journals/corr/abs-2310-10683,DBLP:conf/emnlp/ChenY23}, and graphs~\cite{DBLP:conf/ccs/Chen000H022,wang2023inductive}. These methods promise a scalable solution to comply with privacy regulations while maintaining model utility. 
% Furthermore, beyond legal compliance, unlearning techniques offer benefits in improving model robustness, correcting poisoned or biased data, and enabling dynamic model updates in real-world applications. 
Beyond legal compliance, unlearning improves model robustness, removes poisoned or biased data, and supports dynamic updates.
However, unlearning on knowledge graph (KG) has not yet been extensively explored. Unlike unstructured data, KGs consist of interdependent triples that propagate influence through graph structure, posing unique challenges to unlearning. KGs store and organize open and private information across industries, examples include Wikidata and Google Knowledge Graph.
Providing unlearning interfaces for KGs is crucial, as sensitive data (e.g., personal healthcare information) can make the continued use of KG models trained on deleted data illegal~\cite{DBLP:journals/clsr/VillarongaKL18, ubaydullayeva2023artificial}. 

Despite the advancements in machine unlearning and graph unlearning~\cite{DBLP:conf/acl/JangYYCLLS23,DBLP:conf/acl/WangCYZWY23,DBLP:conf/acl/IsonumaT24,DBLP:conf/iclr/ChengDHAZ23, DBLP:journals/corr/abs-2206-09140,xiao2025knowledge,xiao2025right}, applying existing graph unlearning methods directly to large-scale KG models presents computational challenges. The majority of existing approximate unlearning algorithms rely on computing the influence of deleted data on the trained model. To compute the influence of deleted data, constructing a computational graph (gradient-flow graph) is required. The operation over the computational graph often accounts for the largest portion of a KG model's computational load, mainly determined by the size of the KG, presenting computational challenges upon frequent deletion requests on large-scale KGs. Large-scale KGs are commonplace in real-world applications. For example, Google KG covers 500 billion facts (i.e., triplets) on 5 billion entities. This challenge poses a major hurdle to KG unlearning.

In this paper, we propose an efficient KG unlearning method without constructing the gradient-flow graph. Specifically, we start with computing the sensitivity of a KG model's parameters to the removal of a single data point with the product of
the Hessian matrix inverse and the first-order gradients~\cite{DBLP:conf/icml/KohL17,wu2023gif,koh2017understanding} .
To address the computational
challenges of the inverse-Hessian vector products (IHVPs) in the influence function,  we first decompose a large matrix into smaller matrices to reduce the memory cost. Then we introduce the Fisher information matrix~\cite{ly2017tutorial} to approximate IHVPs. In the probabilistic view, the Fisher information matrices of the model's conditional distribution are equivalent to Hessian matrices. With the approximation, we can express the influence function with the inverse of the empirical Fisher matrices. To further reduce the cost of computing gradient vectors required in IHVPs, we introduce zeroth-order optimization~\cite{chen2017zoo} to approximate the first-order gradients for KG unlearning. Instead of computing the gradients based on the loss value, we use the model output itself (i.e. zeroth-order information) to approximate the values of first-order gradients.
This approach is particularly advantageous in scenarios where gradient information is expensive to compute, such as computing gradients on KG embeddings.

Figure~\ref{fig: workflow} shows the computational flow of (1) traditional influence function (KGIF) without approximation; (2) approximating IHVPs with the Fisher matrices; and (3) approximating IHVPs with the Zeroth-order information. Note that our method can avoid high computational costs over the gradient flow (the computational graph) by approximating with the Zeroth-order information. We verify the efficiency and efficacy of the proposed unlearning method by approximation in the experiment. 

We investigate three KG unlearning tasks: triplets unlearning, entity unlearning, and relation unlearning. Considering that the majority of applications on KG such as question answering, knowledge
acquisition and recommendation rely heavily on KG embedding~\cite{DBLP:journals/tnn/JiPCMY22}, we aim to remove the influence of deleted triplets, entities, and relations from the trained KG embedding models upon request. Different from existing graph unlearning methods developed for graph neural network (GNN) models and classification tasks~\cite{wu2020comprehensive, DBLP:conf/iclr/KipfW17}, our KG unlearning method optimizes a directional scoring loss to learn representations of entities and relations in triplets~\cite{dai2020survey}.

Our high-level contributions are as follows:
\begin{itemize}
    \item We investigate KG unlearning and use an influence function to forget triplets, entities, and relations.
    \item We introduce the Fisher matrices to reduce the computational cost of inverse Hessian vector products.
    \item We further use zeroth-order information to approximate the first-order gradient and reduce the computational cost over the gradient flow.
    \item We experimentally demonstrate that our approximating method can significantly reduce the run time and memory cost. At the same time, our unlearning method outperforms the state-of-the-art graph unlearning baselines.
    
\end{itemize}

\section{Preliminary}
Let $G=(E, R, T)$ be a KG, where $E$ and $R$ are the sets of entities and relations in the KG. We use $T$ to denote the set of triples, each of which is a triplet $(e_h, r, e_t)$, including the head entity $e_h \in E$, the tail entity $e_t \in E$ and the relation $r\in R$ between $e_h$ and $e_t$. In this paper, we explore KG unlearning with embedding models, because the majority of applications on KG such as question answering, Knowledge
acquisition, and recommendation rely heavily on KG embedding according to a survey~\cite{DBLP:journals/tnn/JiPCMY22}. A KG embedding model $M(G)$ trained on $G$ associates each entity and relation with a vector in an embedding space.

The user can request to delete (1) a subset of triples $T_d$; (2) a subset of relations $R_d$; (3)a subset of entities $E_d$. Denote $G/T_d$ as the new KG where $T_d$ has been removed from $G$. The naive unlearning is to retrain a new model $M(G/T_d)$ on the remaining data $G/T_d$ from scratch. Similarly, $M(G/E_d)$ and $M(G/R_d)$ represent the new models for entity unlearning and relation unlearning. To resolve the computational issue of retraining, the goal of this paper is to develop an efficient unlearning algorithm to directly eliminate the effects of deleted data. Ideally, the unlearning algorithm will result in a new KG model with a similar performance as the retrained one. Considering that a KG is stored as triplets, we removed all triplets including the deleted entity or relation. The schema of the KG enables us to query specific triplets (e.g., foaf: Person, rdf: Is a friend of, foaf: Person), entities, or relations to address privacy or concerns from users.

\section{Related Works}
\subsection{Machine Unlearning}
% \noindent\textbf{Machine Unlearning.}
Given the guidelines such as the Right to be Forgotten, machine unlearning methods aim to remove the impact of deleted data on the trained model~\cite{sekhari2021remember, nguyen2022survey,DBLP:conf/acl/WangCYZWY23}. While re-training from scratch on the retraining dataset is the intuitive method, the time and computational costs are not practical in applications. To alleviate the issue, some research efforts have been proposed to unlearn data efficiently. Existing unlearning methods mainly include exact unlearning and approximate unlearning. Exact unlearning aims to learn an unlearning model with the same performance as the retrained one~\cite{DBLP:conf/aistats/LiWC21,DBLP:journals/corr/abs-2106-15093,DBLP:conf/icml/BrophyL21,DBLP:conf/nips/GinartGVZ19,cao2015towards,DBLP:conf/colt/UllahM0RA21,DBLP:conf/www/Chen0ZD22,DBLP:conf/ccs/Chen000HZ21}. For example, SISA~\cite{bourtoule2021machine} divides the dataset into several shards, trains multiple sub-models and establishes an integrated model to reduce the time cost. Approximate unlearning methods try to update the
parameters of the trained model closer to the retraining one through the approximation~\cite{DBLP:conf/aaai/GravesNG21,DBLP:conf/aaai/MarchantRA22,DBLP:conf/icml/WuDD20,DBLP:conf/alt/Neel0S21,DBLP:conf/nips/GinartGVZ19,DBLP:conf/aaai/FosterSB24, DBLP:conf/aaai/ChundawatTMK23}. Currently, there is a new trend to unlearn the specific concepts which are learned in complex data in large language models (LLMs)~\cite{DBLP:journals/corr/abs-2310-10683,DBLP:journals/corr/abs-2402-08787}.

\subsection{Graph Unlearning}
% \noindent\textbf{Graph Unlearning.} 
Different from general unlearning independent instances, deleting nodes and edges on a graph will have effects on other connected nodes and edges. Graph unlearning aims to approximate updates on the embeddings of surrounding nodes or edges~\cite{wu2023gif, cheng2023gnndelete, wu2023certified,DBLP:conf/kdd/DongZ0ZL24}. GraphEraser~\cite{chen2022graph} segment balanced shards reducing the destructiveness of the graph structure. However, the key point of this method lies in the uncontrollability of the number of clusters, hence the efficiency varies in different datasets. GIF~\cite{wu2023gif} uses an additional loss term
of the influenced neighbors to update the graph model. However, it explicitly computes the first derivative of the loss which is a huge computational burden. ~\cite{xiao2025knowledge} analyze the schema patterns in knowledge graph and unlearn data based on cosine similarity of subgraphs within the schema. GNNDELETE~\cite{DBLP:conf/iclr/ChengDHAZ23} studies GNNs unlearning and proposes two standards for optimization: Deleted Edge Consistency and Neighborhood Influence. Some other works leverage KG to assist unlearning in the context of federated Learning~\cite{DBLP:conf/www/ZhuL023} or removing harmful content from LLMs~\cite{DBLP:conf/iclr/MaiJYYHP24,DBLP:conf/emnlp/Feng0LL24}. To the best of the authors' knowledge, our paper is the first work to study unlearning entities or relations on KG models.

\section{Methodology}
\subsection{Knowledge Graphs Influence Functions}
We first define an influence function in the KG embedding model to assess the variation in model parameters when a sample is marginally increased in weight. The benefit of our influence function is that it is universal for all three types of unlearning tasks (triplet unlearning, entity unlearning, and relation unlearning) since all three tasks land on deleting triplets for a KG and the essence is to measure the impact on the KG embeddings. This is different from general graph unlearning where the optimization target variies across different tasks (e.g., deleted edge consistency~\cite{DBLP:conf/iclr/ChengDHAZ23}, classification, and link prediction).

 Without loss of generality, a KG embedding model determines whether there is a specific type of relation between any two entities~\cite{dai2020survey}. Denote $\theta$ as the parameters in the KG embedding model $M(G)$. Let $f(e_h, r, e_t;\theta)$ be the score of $(e_h, r, e_t)$ computed with $\theta$ and the embeddings. Given a positive training set of triplets $(e_h, r, e_t)\in T^+$ and a negative sample set $(e_h', r', e_t'; \theta)\in T^-$, an example of the loss function (TransH~\cite{wang2014knowledge}) can be written as:

\begin{equation}\label{eq:loss2}
    \mathcal{L} = \sum_{T^+} \sum_{T^-} f(e_h, r, e_t; \theta) - f(e_h', r', e_t'; \theta) + \gamma,
\end{equation}
where $\gamma$ is the margin separating
positive and negative triplets. The score $f(\cdot)$ is expected to be lower for a fact triplet and higher for an incorrect triplet.

We remark that our unlearning method is model-agnostic and works for any KG embedding models trained based on a loss function. Given any model trained with a loss function $\mathcal{L}$, the model parameters $\theta$ can be optimized by minimizing:
% \begin{equation}\label{eq:optimal_parameter}
%     \hat{\theta}= \underset{\theta}{\arg\min} \mathcal{L}
% \end{equation}
\begin{equation}\label{eq:optimal_parameter}
    \hat{\theta}= \arg\min\mathcal{L}
\end{equation}
where $\hat{\theta}$ be the optimized parameters trained on $G$. Suppose that $T_d$ is the set of triplets to be deleted. To estimate the influence of $T_d$, we add a small perturbation caused by data removal (e.g., delete a small set of triplets) to the loss:
\begin{equation}
\label{eq:perturbation}
    \Tilde{\theta}=\arg\min(\mathcal{L}_{\theta_0}+\epsilon\mathcal{L}_{T_d}),
\end{equation}
% \begin{equation}
% \label{eq:perturbation}
%     \Tilde{\theta}=\underset{\theta}{\arg\min}(\mathcal{L}_{\theta_0}+\epsilon\mathcal{L}_{T_d}),
% \end{equation}
where $\epsilon$ is a scalar. For simplicity, we use $\mathcal{L}_{T_d}$ to represent the loss value concerning all triplets in $T_d$. For example, $\mathcal{L}_{T_d}=\sum\limits_{T_d} \sum\limits_{{T^-}} f(e_h, r, \\ e_t; \theta) - f(e_h', r', e_t'; \theta) + \gamma$ in TransH with new negative samples $T^-$ for $T_d$. Note that $\Tilde{\theta}$ is optimized parameters for Eq.~(\ref{eq:perturbation}). Therefore, we have the first-order optimal conditions as follows: 
\begin{equation}\label{eq:first_order_optimal}
    0=\nabla_{\hat{\theta}}\mathcal{L}, \text{ }0=\nabla_{\Tilde{\theta}}\mathcal{L}_0 + \epsilon\nabla_{\Tilde{\theta}}\mathcal{L}_{(T_d)}.
\end{equation}
These two conditions are used to ensure that local minima of the loss function are found under different circumstances.

When the permutation is small enough ($\epsilon \to 0$), the perturbed parameters $\Tilde{\theta}$ will converge towards $\Tilde{\theta}\approx \hat{\theta}$.
We use a first-order Taylor expansion to approximate the behavior of the parameter changes under perturbation:
% \begin{equation}\label{eq:first_order_taylor_epansion}
% \begin{split}
%     0 \approx &(\Tilde{\theta}-\hat{\theta})\left(\nabla^2_{\hat{\theta}} \mathcal{L} + \epsilon \nabla^2_{\hat{\theta}} \mathcal{L}_{T_d}\right) \\
%     &+ \left(\nabla_{\hat{\theta}} \mathcal{L}_0+\epsilon \nabla_{\hat{\theta}} \mathcal{L}_{T_d} \right)
%     \end{split}
% \end{equation}

\begin{equation}\label{eq:first_order_taylor_epansion}
    0 \approx (\Tilde{\theta}-\hat{\theta})\left(\nabla^2_{\hat{\theta}} \mathcal{L} + \epsilon \nabla^2_{\hat{\theta}} \mathcal{L}_{T_d}\right) + \left(\nabla_{\hat{\theta}} \mathcal{L}_0+\epsilon \nabla_{\hat{\theta}} \mathcal{L}_{T_d} \right)
\end{equation}
% Our target is to estimate the difference between $\hat{\theta}$ and $\hat{\theta}_{G/T_d}$ where $\hat{\theta}_{G/T_d}$ are the model parameters of the KG model after unlearning. 

% \begin{equation}
%     \hat{\theta}-\hat{\theta}_{G/T_d} \approx H_{\hat{\theta}}^{-1} \nabla_{\hat{\theta}} \mathcal{L}(\hat{\theta}),
% \end{equation}
% where $\hat{\theta}_{G/T_d}$ are the model parameters of the KG model after unlearning. 

% We use $\nabla_{\hat{\theta}}$ to represent the gradient of the loss function and $H_{\hat{\theta}}^{-1}$ is the inverse of hessian matrix of the loss function. 

Recall that our task is to eliminate the impact of $T_d$ on the training process of $\theta$. Therefore, we assign the value of $\epsilon$ as -1 to cancel out the effect of $T_d$. Recall that $0=\nabla_{\hat{\theta}}\mathcal{L}$ from Eq.~(\ref{eq:first_order_optimal}) and the second-order derivatives $\nabla^2_{\hat{\theta}}$ can be omitted because $T_d$ is a small part of $G$. For the purpose of simplicity, we use $H_{\hat{\theta}}$ to represent the Hessian Matrix $\nabla^2_{\hat{\theta}} \mathcal{L}$ and $v$ to denote the first-order gradient $\nabla_{\hat{\theta}} \mathcal{L}_{T_d}$. Then when $\epsilon=-1$, we have:
% \vspace{-1mm}
\begin{equation}\label{eq:parameter_change}
    \Tilde{\theta}-\hat{\theta} \approx H_{\hat{\theta}}^{-1} v.
    % \vspace{-1mm}
\end{equation}

Intuitively, deleting an entity or a relation will have an impact on the embeddings of the entities and relationships in the neighborhood. The impact on these entities far from the deleted triplets on the topology is marginal.  Therefore, updating embeddings by query entities and relations within the K-hop neighborhood around the deleted component can reduce the cost of updating embeddings of large-scale KG.
% \begin{equation}\label{eq:k-hop_entity}
%     \mathcal{N}_k (e_i) = \{ e_j \mid 1 \leq \text{SPD}(e_j, e_i) \leq k \}
% \end{equation}
% \begin{equation}\label{eq:k-hop_relation}
%     \mathcal{N}_k (r_i) = \mathcal{N}_k (e'_1) \cup \mathcal{N}_k (e'_2) \cup \{e'_1, e'_2\}
% \end{equation}
% \begin{equation}\label{eq:k-hop_triples}
% \begin{split}
%     \mathcal{S}_d = &\{ (e, r, e') \mid e, e' \in \mathcal{N}_k(e_i) \cup \mathcal{N}_k(e'_1)\\ &\cup \mathcal{N}_k(e'_2), r \in \mathcal{N}_k(r_i) \}
%     \end{split}
% \end{equation}

% According to the influencing regime in Equation \ref{eq:k-hop_triples}, we can clarify the results and scope of $H_{\hat{\theta}}^{-1} \nabla_{\hat{\theta}} \mathcal{L}(\hat{\theta}_{S_d})$. 

We remark that the largest computational load is to calculate the inverse of the Hessian matrix and the inverse-Hessian
vector products (IHVPs) in Eq.~(\ref{eq:parameter_change}). Specifically, Eq.~(\ref{eq:parameter_change}) entails a computational complexity of \(O(\|\theta\|^3 + n\|\theta\|^2)\) and requires memory allocation of \(O(\|\theta\|^2)\). It is impractical to directly compute Eq.~(\ref{eq:parameter_change}) in real-world applications due to the cost. For example, there are 5 billion entities in Google Knowledge Graph. Suppose the length of the embeddings is 128. There will be 128$\times$5 billion parameters to be updated for entities, which may impede the implementation of the unlearning algorithm in practice.
By leveraging the stochastic estimation method, we can reduce the complexity and memory to \(O(n\|\theta\|^2)\) and \(O(\|\theta\|)\) by iterating the following equation until convergence~\cite{cook1980characterizations}:
\begin{equation}
   H_s^{-1} =(I - H_{\hat{\theta}}) H_{s-1}^{-1} + I,
\end{equation}
where $I$ is the identity matrix. The IHVPs will be updated through the Taylor series expansion:
\begin{equation}\label{eq:gif_iteration}
H_s^{-1}v = v + H_{s-1}^{-1} v - \eta H_{\hat{\theta}} H_{s-1}^{-1} v.
\end{equation}
The coefficient $\eta$ is used to control the convergence condition. Note that $v$ remains unchanged in the iteration, though it is frequently used for computation.  We can store it at the
end of the model training. These calculations can be performed using the PyTorch Autograd Engine.

\subsection{Approximation with the First-order Information}
Although we have reduced the computational complexity through \cite{cook1980characterizations}, computing the gradient and IHVPs still incurs significant computational overhead. As its computational complexity is determined by the gradient size of the knowledge graph model, unlearning on a large scale KG is time-consuming and memory-consuming due to the involved
large computation graphs for BP (back-propagation).  In the experiment, we show that directly applying the influence functions for KG unlearning takes a substantial amount of time and consumes a substantial amount of GPU memory.

Considering that the KG size is usually huge in real applications, computing the product of first-order (gradients) and second-order (Hessian Matrix Inverse) in Eq.~(\ref{eq:parameter_change}) requires the construction of a complete computation graph (gradient flow) to track gradients for parameter updates.

To speed up IHVPs computation and
reduce memory complexity, we introduce Woodbury theorem~\cite{woodbury1950inverting} to allow cheap computation of inverses. Woodbury matrix identity says that the inverse of a rank-k correction of some matrix can be computed by doing a rank-k correction to the inverse of the original matrix. Specifically, it decomposes a large matrix into much smaller matrices, allowing the influence functions to operate on smaller matrices, thereby reducing the computational cost. Formally, the inverse of a large matrix can be computed as: 
\begin{equation}
    (A + U\Lambda^\top)^{-1} = A^{-1} - A^{-1}U(I + \Lambda^\top A^{-1} U)^{-1}\Lambda^\top A^{-1}
\end{equation}
In our setting, we have $H\approx U\Lambda^\top + A $, where (1) $A$ usually represents a fundamental matrix that is invertible, positively definite, or at least non-singular; (2) $U$ usually denotes a group of directions or factors, which will be added to the basic matrix $A$ to modify or adjust some characteristics of $A$, such as updated parameters in KGs models, (3) $\Lambda$ is used in conjunction with $U$ to indicate how modifications to $A$ are balanced across different dimensions or directions. Note that decomposition provides a thorough understanding of how the change in each dimension of model parameters will impact prediction outcomes.

% \textbf{Fisher} is a scrubbing process that eliminates information from the trained weights without requiring access to the original training data or the need to retrain the whole network~\cite{golatkar2020eternal}. 

Next, we introduce Fisher Information Matrix (FIM)~\cite{ly2017tutorial} to efficiently estimate the inverse Hessian instead of using direct inversion techniques. FIM offers a powerful method to compute a
faithful and efficient estimate of the inverse Hessian. 

Formally, FIM quantifies the amount of information that an observable random function $\mathcal{L}$ carries about unknown parameters $\theta$ upon which the model distribution depends. The definition of the FIM is given by:
\begin{equation}
    \mathcal{F} = \mathbb{E}\left[\nabla_\theta \log p(\mathcal{L}; \theta) \nabla_\theta \log p(\mathcal{L}; \theta)^\top\right]
\end{equation}
where $\nabla_\theta \log p(\mathcal{L}; \theta)$ is the gradient of the log-likelihood of the model concerning the parameters $\theta$,  $p(\mathcal{L}; \theta)$ is the density function corresponding to the model distribution, and $\mathbb{E}$ denotes the expectation over the distribution of $\mathcal{L}$ parameterized by $\theta$. It can be further written as: $\mathcal{F} = \mathbb{E}_{P_{\theta,\mathcal{L}}} \left[ -\nabla^2 \log p(\theta, \mathcal{L}) \right]$.

Since the probability of different relations between two entities in KG can be considered as a conditional distribution $P_{\mathcal{L}|\theta}$ over the model output (i.e., embeddings), it can be proved that the Fisher and Hessian matrix $H$ are equivalent by expressing $P_{\mathcal{L}|\theta} = Q_\theta P_{\mathcal{L}|\theta} \approx \hat{Q}_\theta P_{\mathcal{L}|\theta}$ under the assumption that the model’s conditional distribution \(P_{\mathcal{L}|\theta}\) matches the conditional distribution of the training data \(\hat{Q}_{\mathcal{L}|\theta}\)~\cite{singh2020woodfisher,ly2017tutorial}. 

Then by combining two approximation methods (i.e., Woodbury theorem and Fisher Information Matrix), the approximation is equivalent to a quasi-Newton approximation~\cite{hazan2016introduction} which can be written as:
\begin{equation}\label{eq:fo_woodfisher}
    H\approx U\Lambda^\top + A \approx \mathcal{F} + A\approx vv^\top + \gamma I
\end{equation}

We can then easily obtain the IHVPs instead of using direct inversion techniques:

\begin{equation}\label{Woodfisher_hvps}
    H_0^{-1}v\approx\gamma^{-1}v+\frac{\gamma^{-2}vv^\top}{1+\gamma^{-1}v^\top v}v
\end{equation}

In Eq.(~\ref{Woodfisher_hvps}), $\gamma > 1$ is a damping term. Recall that we update $H^{-1}$ with $H_s^{-1}v = v + H_{s-1}^{-1} v - H_{\hat{\theta}} [H_{s-1}^{-1}] v$ in each iteration. We have:

\begin{equation}
\label{eq:damp}
\begin{split}
    \quad [H_{1}^{-1}] v =&v+ (\gamma^{-1}v+\frac{\gamma^{-2}vv^\top}{1+\gamma^{-1}v^\top v}v)v\\
    &-\eta H_{\hat{\theta}}(\gamma^{-1}v+\frac{\gamma^{-2}vv^\top}{1+\gamma^{-1}v^\top v}v)v,
    \end{split}
\end{equation}
Iteratively, we can optimize $\theta$ with Eq.~(\ref{eq:parameter_change}). At this point, we have converted the second-order information in Eq.(~\ref{eq:parameter_change}) to first-order information with an approximation. It significantly reduces the computational complexity and the memory cost for KG unlearning.

\subsection{Approximation with the Zeroth-order Information}
Note that we still need to compute the gradient $v$ in Eq.(~\ref{Woodfisher_hvps}) by constructing a computational graph, though the approximation with the first-order information avoids the direct computation of the inverse Hessian matrix $H^{-1}$ and IHVPs. To further reduce the computational cost, we approximate the first-order gradients with zeroth-order information motivated by~\cite{DBLP:conf/nips/MalladiGNDL0A23, DBLP:journals/corr/abs-2402-11592}.

Specifically, we approximate the gradient with the loss function values at a point by leveraging the Taylor series expansion. In general, the central difference is more accurate in providing the derivative of a function than other forward or backward difference approximation methods.
By conducting Second-order Taylor expansion on $\mathcal{L}$, we have the central difference:
\begin{equation}\label{eq:postive}
    \mathcal{L}(\hat{\theta}+\varepsilon)=\mathcal{L}(\hat{\theta})+\varepsilon\mathcal{L}'(\hat{\theta})+\frac{\varepsilon^2}{2}\mathcal{L}''(\hat{\theta})+O(\varepsilon^3)
\end{equation}
\begin{equation}\label{eq:negtive}
    \mathcal{L}(\hat{\theta}-\varepsilon)=\mathcal{L}(\hat{\theta})-\varepsilon\mathcal{L}'(\hat{\theta})+\frac{\varepsilon^2}{2}\mathcal{L}''(\hat{\theta})+O(\varepsilon^3)
\end{equation}
\begin{equation}
\label{eq:central}
    v\approx \frac{\mathcal{L}(\hat{\theta}+\varepsilon)- \mathcal{L}(\hat{\theta}-\varepsilon)}{2\varepsilon}=\mathcal{L'}(\hat{\theta})+O(\varepsilon^2),
\end{equation}
where the error term is $O(\varepsilon^2)$. Note also that the forward difference and backward difference are given by:
\begin{equation}\label{eq:postive2}
    \mathcal{L}(\hat{\theta}+\varepsilon)=\mathcal{L}(\hat{\theta})+\varepsilon\mathcal{L}'(\hat{\theta})+O(\varepsilon^2)
\end{equation}
\begin{equation}
    v\approx \frac{\mathcal{L}(\hat{\theta}+\varepsilon)- \mathcal{L}(\hat{\theta})}{\varepsilon}=\mathcal{L'}(\hat{\theta})+O(\varepsilon)
\end{equation}
\begin{equation}\label{eq:negtive2}
    \mathcal{L}(\hat{\theta}-\varepsilon)=\mathcal{L}(\hat{\theta})-\varepsilon\mathcal{L}'(\hat{\theta})+O(\varepsilon^2)
\end{equation}
\begin{equation}
    v\approx \frac{\mathcal{L}(\hat{\theta})- \mathcal{L}(\hat{\theta}-\varepsilon)}{\varepsilon}=\mathcal{L'}(\hat{\theta})+O(\varepsilon)
\end{equation}
Where the error term is $O(\varepsilon)$.

When $\varepsilon$ is small ($<1$), $\varepsilon^2$ is much smaller than $\varepsilon$. Therefore, 
the gradient approximated by the central difference in Eq.~(\ref{eq:central}) is more accurate than the forward difference $\frac{\mathcal{L}(\hat{\theta}+\varepsilon)-\mathcal{L}(\hat{\theta})}{\varepsilon}$ and the backward difference $\frac{\mathcal{L}(\hat{\theta})-\mathcal{L}(\hat{\theta}-\varepsilon)}{\varepsilon}$.
Then we estimate the first-order information with:
\begin{equation}
\label{eq:zeroth-order}
    \mathcal{L}'(\hat{\theta})\approx\frac{\mathcal{L}(\hat{\theta}+\varepsilon)-\mathcal{L}(\hat{\theta}-\varepsilon)}{2\varepsilon},
\end{equation}
which can avoid the construction of the computational graphs to compute the gradients. By replacing the gradient $v$ in Eq.(~\ref{eq:parameter_change}) with the zeroth-order information in Eq.~(\ref{eq:zeroth-order}), our unlearning algorithm only utilizes the zeroth-order information without computing the gradients and the inverse Hessian of the parameters.

\noindent\textbf{Discussion.} In summary, our unlearning algorithm updates the KG model with the IHVPs approximation and the gradient approximation based on deleted triplets. While all existing approximate unlearning algorithms are estimations of retraining, our approximation is efficient and close to retraining, which can be verified in the experiment. Our algorithm has three important hyperparameters: iterations $s$, damping term $\gamma$, scaling down term, and noise term $\varepsilon$. The value of $s$ decides the number of iterations of IHVPs. The damping term $\gamma$ is used to reduce the amplitude of parameter variation estimation. The scaling-down term $\eta$ is used as the denominator of the parameter change estimation vector to control the magnitude of parameter changes. The noise term $\varepsilon$ is used to perturb the intensity of parameter estimation gradients. We will investigate their impacts in the experiment.

\definecolor{lightblue}{RGB}{217,233,243}
\definecolor{lightred}{RGB}{239,200,196}
\definecolor{lightgray}{gray}{0.9}
\begin{table*}[t]
    \centering
    \resizebox{1.00\linewidth}{!}{
        \begin{tabular}{|l|l|>{\columncolor[gray]{0.95}}c>{\columncolor[gray]{0.95}}c|ccc|>{\columncolor[gray]{0.95}}c>{\columncolor[gray]{0.95}}c|ccc|} 
        \hline
        
        \multirow{2}{*}{Model} &\multirow{2}{*}{Method}&\multicolumn{5}{c|}{FB15K237}&\multicolumn{5}{c|}{YAGO3}\\
        & &Time&Memory & MRR & Hit@3 & Hit@1 & Time & Memory& MRR & Hit@3 & Hit@1\\ 
        \hline
        \multirow{7}{*}{RotatE} & Train & 1666s & 932MB & 0.3165  & 0.3506 & 0.2251 & 11049s & 4493MB & 0.3381 & 0.3846 & 0.2356 \\
        \cline{2-12}
        \rowcolor{lightred}
        \cellcolor{white} & Retrain & 1508s & 849MB & 0.2806  & 0.3135 & 0.1880 & 9619s & 4103MB & 0.2953 & 0.3444 & 0.1912\\
        \cline{2-12}
        & GNNDelete & 347s & 804MB & 0.2433 & 0.2720 & 0.1522 & 1441s & \textbf{3278MB}& 0.2499 & 0.2896 & 0.1492\\
        \cline{2-12}
        & CertifiedGU & 224s & 9336MB & 0.2627  & 0.2938 & 0.1658 & 214s & 38653MB & 0.2623 & 0.3113 & 0.1547 \\
        \cline{2-12}
        & KGSchema& 94s & 862MB & 0.2548 & 0.2911 & 0.1579 & 1027s& 1749MB& 0.2612 & 0.3079 & 0.1478 \\
        \cline{2-12}
        & GIF & 9.3s & 6379MB & \textbf{0.2713}   & 0.3041 & \textbf{0.1769} & 12.7s & 25954MB & 0.2656 & 0.3171 & 0.1558\\
        \cline{2-12}
        & WF-KGIF (Our) & 0.55s & 1958MB & 0.2712 & 0.3042 & 0.1768& 0.71s & 8851MB & 0.2660  & 0.3170 & 0.1559 \\
        \cline{2-12}
        \rowcolor{lightblue} 
        \cellcolor{white} & ZeroFisher (Our) & \textbf{0.48s} & \textbf{771MB} & 0.2710  & \textbf{0.3042} & 0.1757 & \textbf{0.62s} & 3357MB & \textbf{0.2664}  & \textbf{0.3176} & \textbf{0.1563}\\
        \hline
        \multirow{7}{*}{TransD} & Train & 1702s & 893MB & 0.2928  & 0.3314 & 0.1971 & 8543s & 3642MB & 0.3197  & 0.4542 & 0.2796\\
        \cline{2-12}
        \rowcolor{lightred}         
        \cellcolor{white} & Retrain & 1545s & 806MB & 0.2578  & 0.2936 & 0.1612 & 8236s & 3287MB & 0.3128 & 0.3765 & 0.1983\\
        \cline{2-12}
        & GNNDelete & 401s & 806MB & 0.2471  & 0.2785 & 0.1544 & 1884s & 3287MB & 0.2637 & 0.3110 & 0.1563\\
        \cline{2-12}
        & CertifiedGU & 613s & 30490MB & \textbf{0.2530}  & 0.2857 & \textbf{0.1562} & - & OOM & - & - & -\\
        \cline{2-12}
        & KGSchema & 156s & 846MB & 0.2476 & 0.2799 & 0.1576 & 1273S& 1276MB& 0.2679 & 0.3443 & 0.1577 \\
        \cline{2-12}
        & GIF & 6.7s & 21274MB & 0.2480  & 0.2819 & 0.1503 & 17.6s & 85524MB & 0.2820 & 0.3515 & 0.1569\\
        \cline{2-12}
        & WF-KGIF (Our) & 0.51s & 4299MB & 0.2492  & 0.2830 & 0.1517 & 1.67s & 18724MB & 0.2820 & 0.3547 & \textbf{0.1615}\\
        \cline{2-12}
        \rowcolor{lightblue} 
        \cellcolor{white} & ZeroFisher (Our) & \textbf{0.49s} & \textbf{645MB} & \textbf{0.2530} & \textbf{0.2858} & \textbf{0.1562} & \textbf{1.39s} & \textbf{2953MB} & \textbf{0.2864}  & \textbf{0.3583} & 0.1593\\    
        \hline
        \multirow{7}{*}{TransH} & Train & 1804s & 773MB & 0.2951  & 0.3315 & 0.2008 & 8421s & 3118MB & 0.4212  & 0.4851 & 0.3108\\
        \cline{2-12}
        \rowcolor{lightred}         
        \cellcolor{white} & Retrain & 1591s & 696MB & 0.2471  & 0.2873 & 0.1447 & 7791s & 2808MB & 0.3212  & 0.3869 & 0.2075\\
        \cline{2-12}
        & GNNDelete & 161s & 741MB & 0.2283  & 0.2551 & \textbf{0.1421} & 1867s & 3004MB & 0.2659  & 0.3115 & 0.1618\\
        \cline{2-12}
        & CertifiedGU & 596s & 29458MB & 0.2536  & 0.2800 & 0.1577 & - & OOM & - & - & -  \\
        \cline{2-12}
        & KGSchema & 77s & 734MB & 0.2533 & 0.2813 & 0.1497 & 736s& 798MB & 0.2837 & 0.3478 & 0.1633 \\
        \cline{2-12}
        & GIF & 6.5s & 20460MB & 0.2444  & 0.2757 & 0.1491 & 16s & 81935MB & 0.2926  & 0.3633 & \textbf{0.1689}\\
        \cline{2-12}
        & WF-KGIF (Our) & 0.46s & 4023MB & \textbf{0.2481} &  0.2800 & 0.1521 & 1.3s & 16670MB & 0.2921 & 0.3653 & 0.1680  \\
        \cline{2-12}
        \rowcolor{lightblue} 
        \cellcolor{white} & ZeroFisher (Our) & \textbf{0.43s} & \textbf{524MB} & 0.2505 & \textbf{0.2825} & 0.1544 & \textbf{1.1s} & \textbf{2322MB} & \textbf{0.2947}  & \textbf{0.3691} & 0.1659\\
        \hline
        \end{tabular}
    }
    \vspace{1mm}
    \caption{Performance on FB15K237 and YAGO3.}
    \vspace{-3mm}
\label{table: results}
\end{table*}

\begin{figure*}[!h]\label{image:Ablation}
\centering
\begin{subfigure}[b]{0.24\textwidth}
        \includegraphics[width=\textwidth]{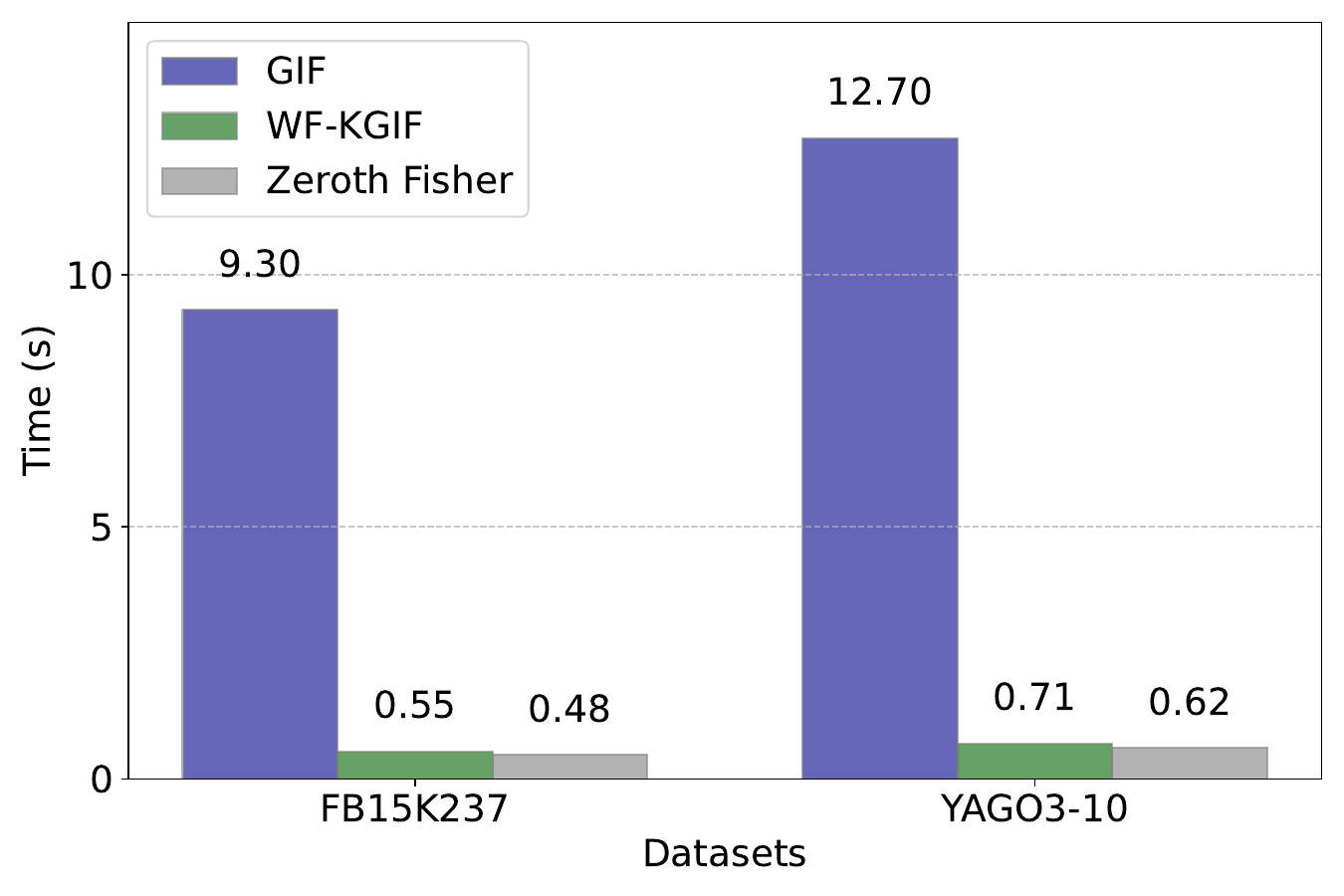}
        \caption{Time Ablation}
        \label{fig:RotatE_Time_Ablation}
    \end{subfigure}
\begin{subfigure}[b]{0.24\textwidth}
        \includegraphics[width=\textwidth]{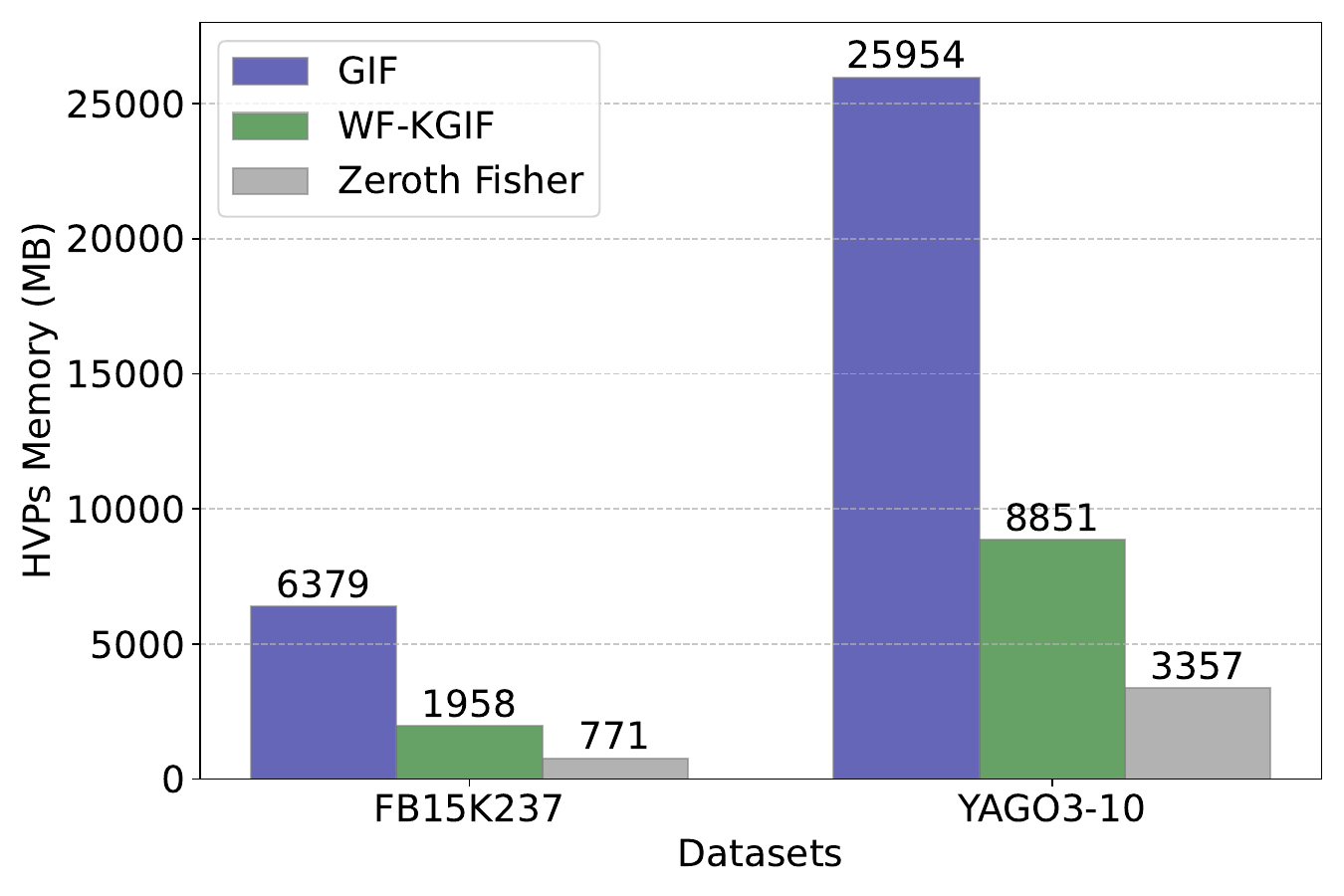}
        \caption{IHVPs Ablation}
        \label{fig:RotatE_HVPs_Ablation}
    \end{subfigure}
\begin{subfigure}[b]{0.24\textwidth}
        \includegraphics[width=\textwidth]{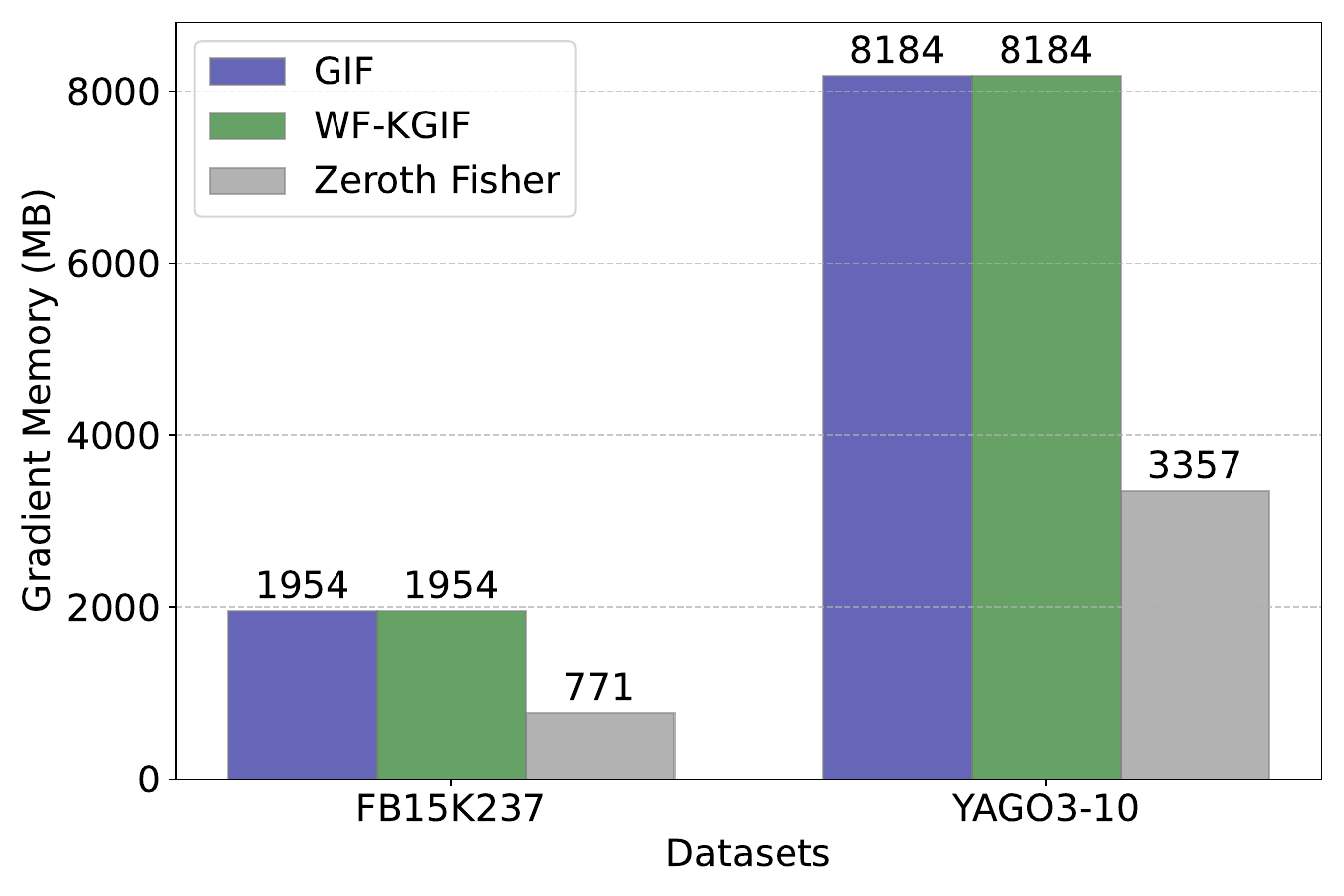}
        \caption{Gradient Ablation}
        \label{fig:RotatE_Gradient_Ablation}
    \end{subfigure}
\begin{subfigure}[b]{0.24\textwidth}
        \includegraphics[width=\textwidth]{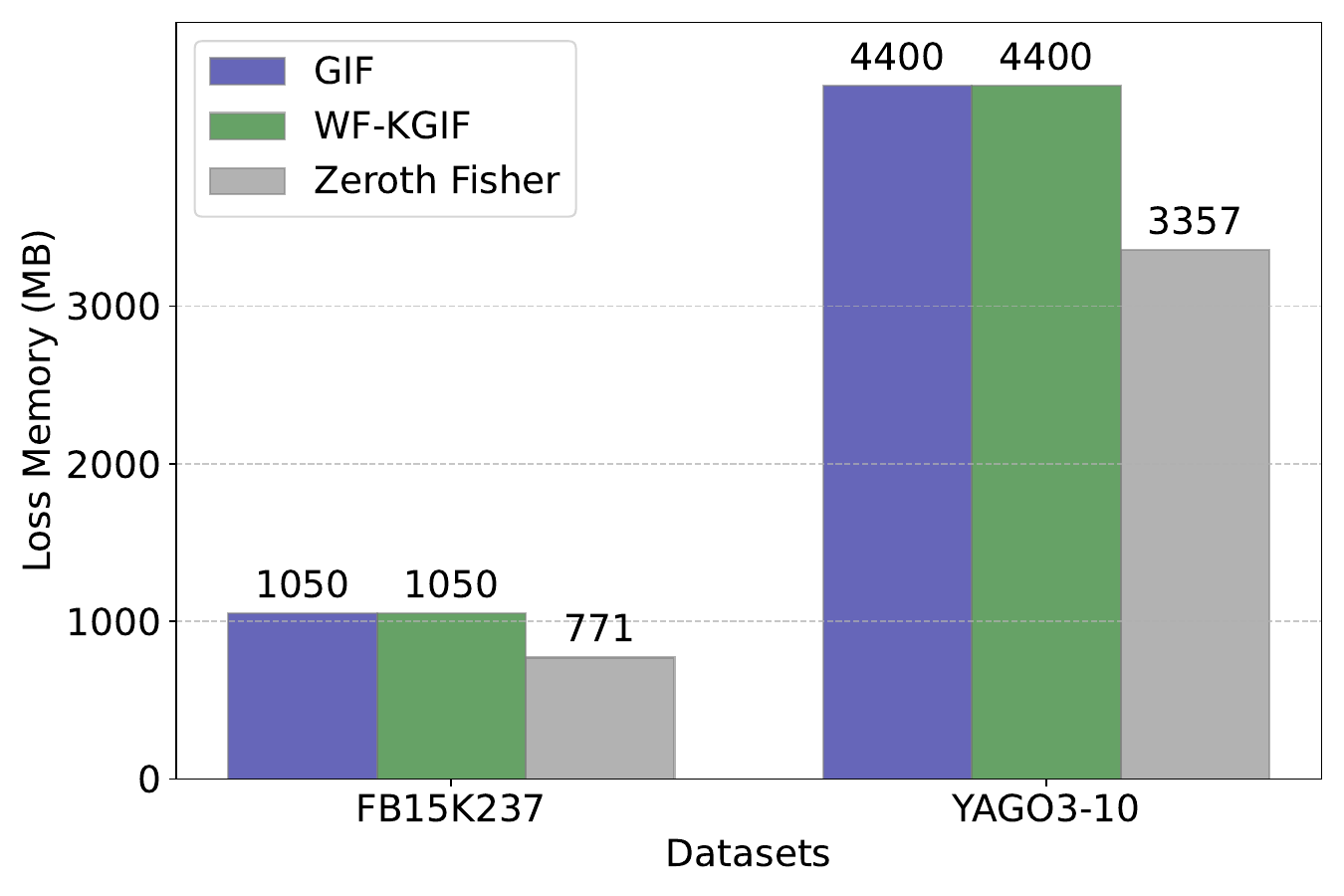}
        \caption{Loss Ablation}
        \label{fig:RotatE_Loss_Ablation}
    \end{subfigure}
    \vspace{-2mm}
    \caption{Ablation study on RotatE}
    \vspace{-2mm}
    \label{fig:ablation}
\end{figure*}

\section{Experiments}

\subsection{Experimental Setup}
In the experiment, we introduce three popular KG models RotatE~\cite{sun2019rotate}, TransD~\cite{ji2015knowledge}, TransH~\cite{wang2014knowledge}. Note that our unlearning method can be applied to any KG embedding method based on a smooth loss function. We trained the KG embedding models on benchmark datasets: FB15K237 and YAGO3-10. We randomly select 5\% triplets, entities, and relations to conduct triplet unlearning, entity unlearning, and relations unlearning, respectively. To verify the superiority of our proposed algorithm, we introduce the state-of-the-art baselines for comparison: GNNDelete~\cite{cheng2023gnndelete}, Certified Graph Unlearning~\cite{chien2022efficient}, GIF~\cite{wu2023gif}. Note that all existing graph unlearning can not be directly used for KG unlearning. We modify the loss function in the baseline to adapt KG models. 
 The number of epochs is set to 1000. We use the Adam optimizer~\cite{kingma2014adam} for RotatE and the SGD optimizer for the other two models (TransD and TransH). The default value of the iteration term is 100, the damping term is 1.00, the scaling down term is 10, and the noise is 1e-5.  The number of iterations for influence functions is set as 100 by default. All experiments are run on two Nvidia L40S GPUs. 
 
\noindent\textbf{Evaluation.} 
To verify the efficiency of our unlearning algorithm, we report the runtime of retraining and all unlearning algorithms. We also recorded the peak graphics memory usage in the unlearning process. These metrics collectively provide insights into the
resources needed for unlearning, and assessing its feasibility in practical scenarios.

\begin{figure*}[t]\label{image:ErrorBar}
\centering
\begin{subfigure}[b]{0.32\textwidth}
        \includegraphics[width=\textwidth]{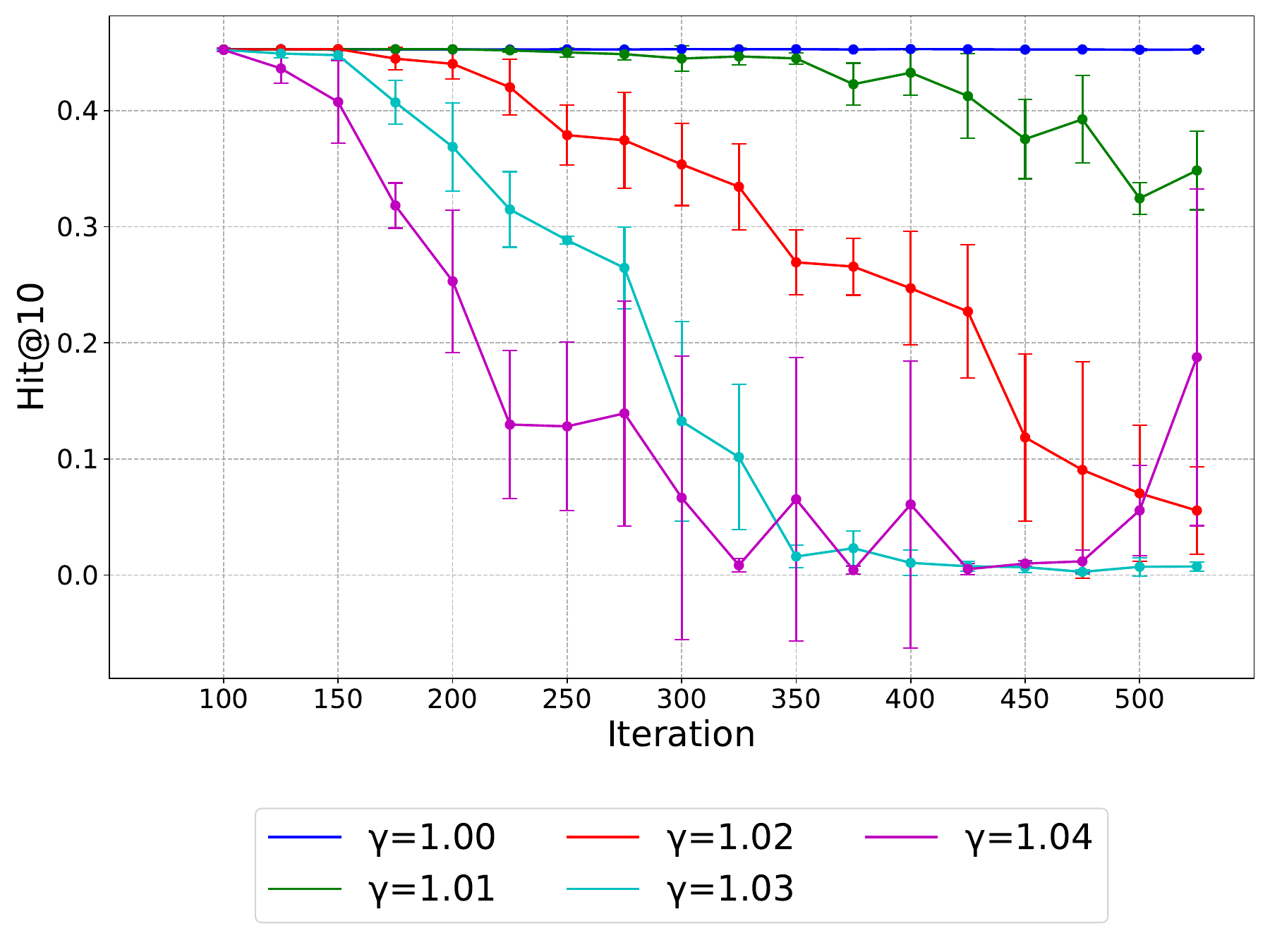}
        \caption{$\gamma$ Term}
        \label{fig:TransH_Iteration-Damp}
    \end{subfigure}
    \begin{subfigure}[b]{0.32\textwidth}
        \includegraphics[width=\textwidth]{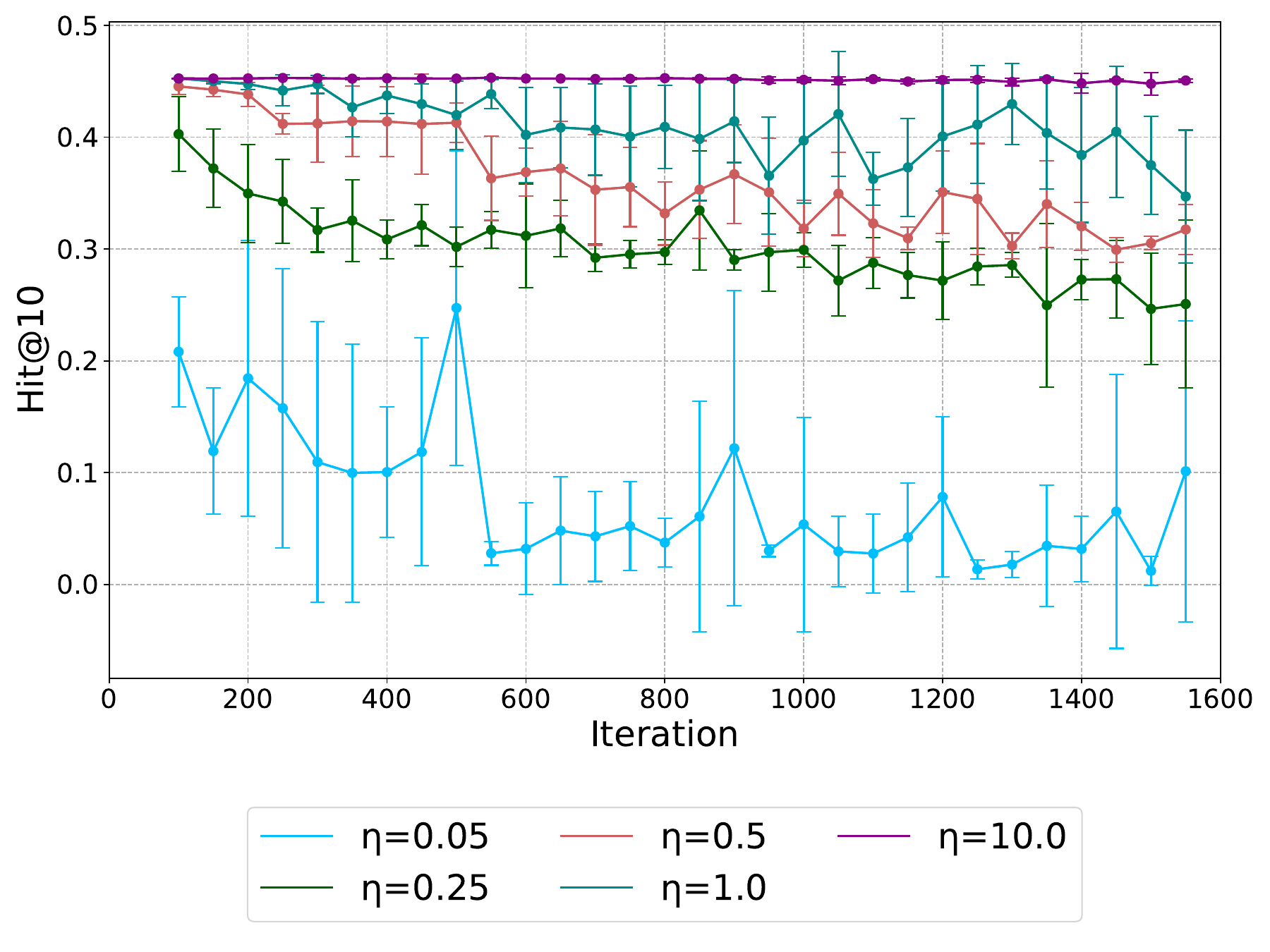}
        \caption{$\eta$ Term}
        \label{fig:TransH_Iteration-Epsilon}
    \end{subfigure}
    \begin{subfigure}[b]{0.32\textwidth}
        \includegraphics[width=\textwidth]{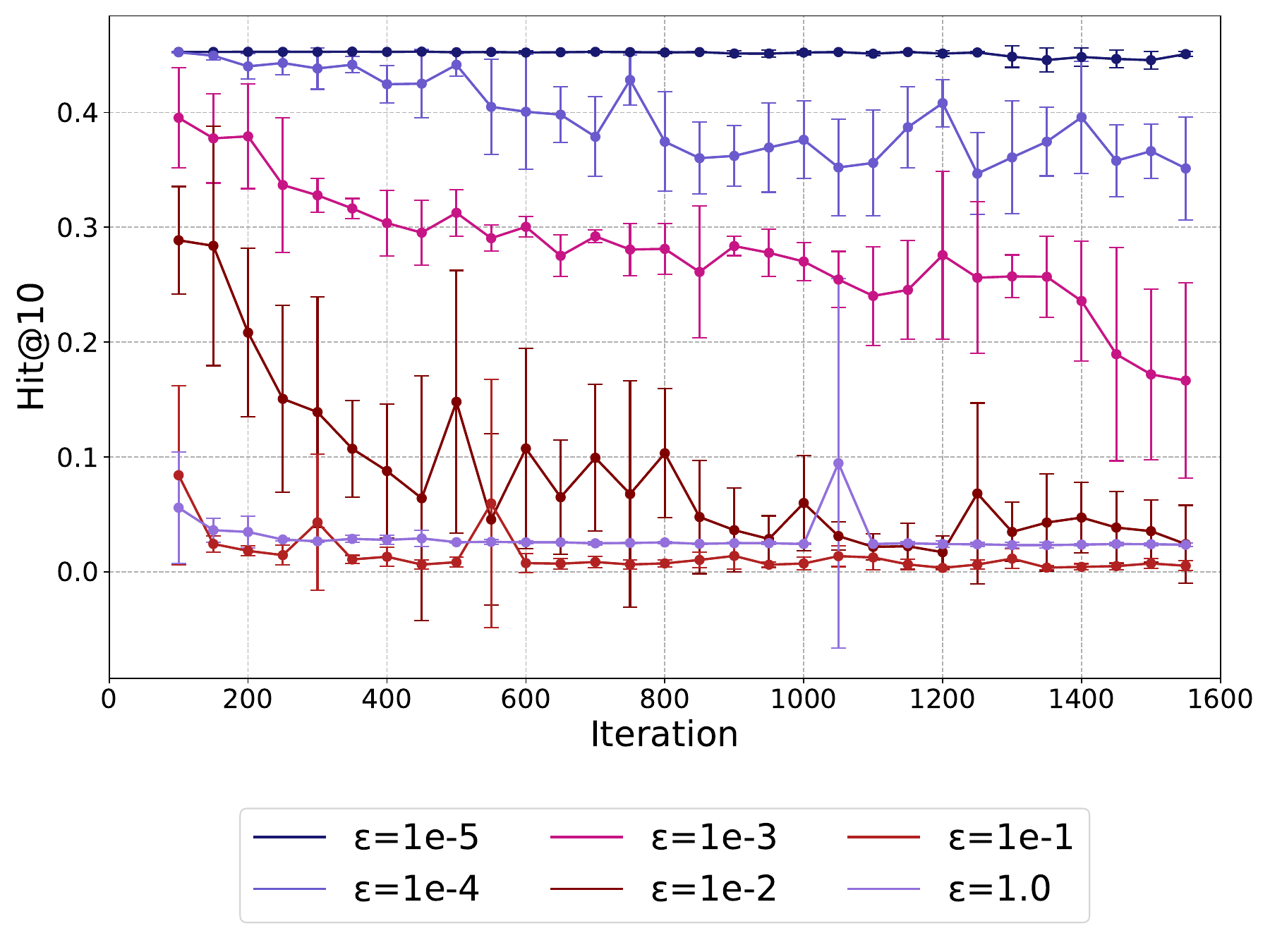}
        \caption{$\varepsilon$ Term}
        \label{fig:TransH_Iteration-Scale}
    \end{subfigure}
    \vspace{-3mm}
    \caption{Varying parameters}
    \vspace{-3mm}
    \label{fig:Varying}
\end{figure*}

\begin{figure*}[t]\label{image:Embedding}
\centering
\begin{subfigure}[b]{0.19\textwidth}
        \includegraphics[width=\textwidth]{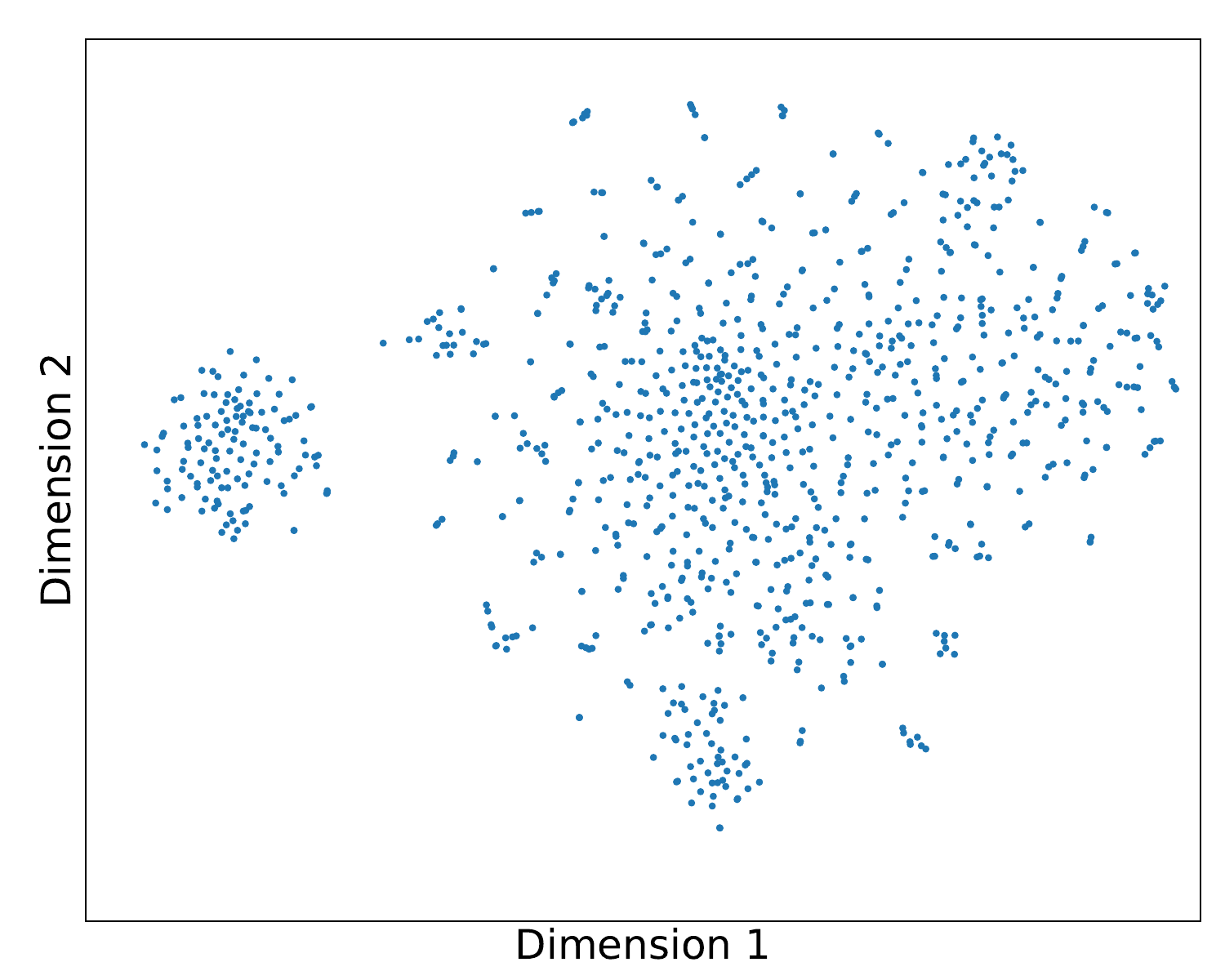}
        \caption{Original}
        \label{fig:original}
    \end{subfigure}
    \begin{subfigure}[b]{0.19\textwidth}
        \includegraphics[width=\textwidth]{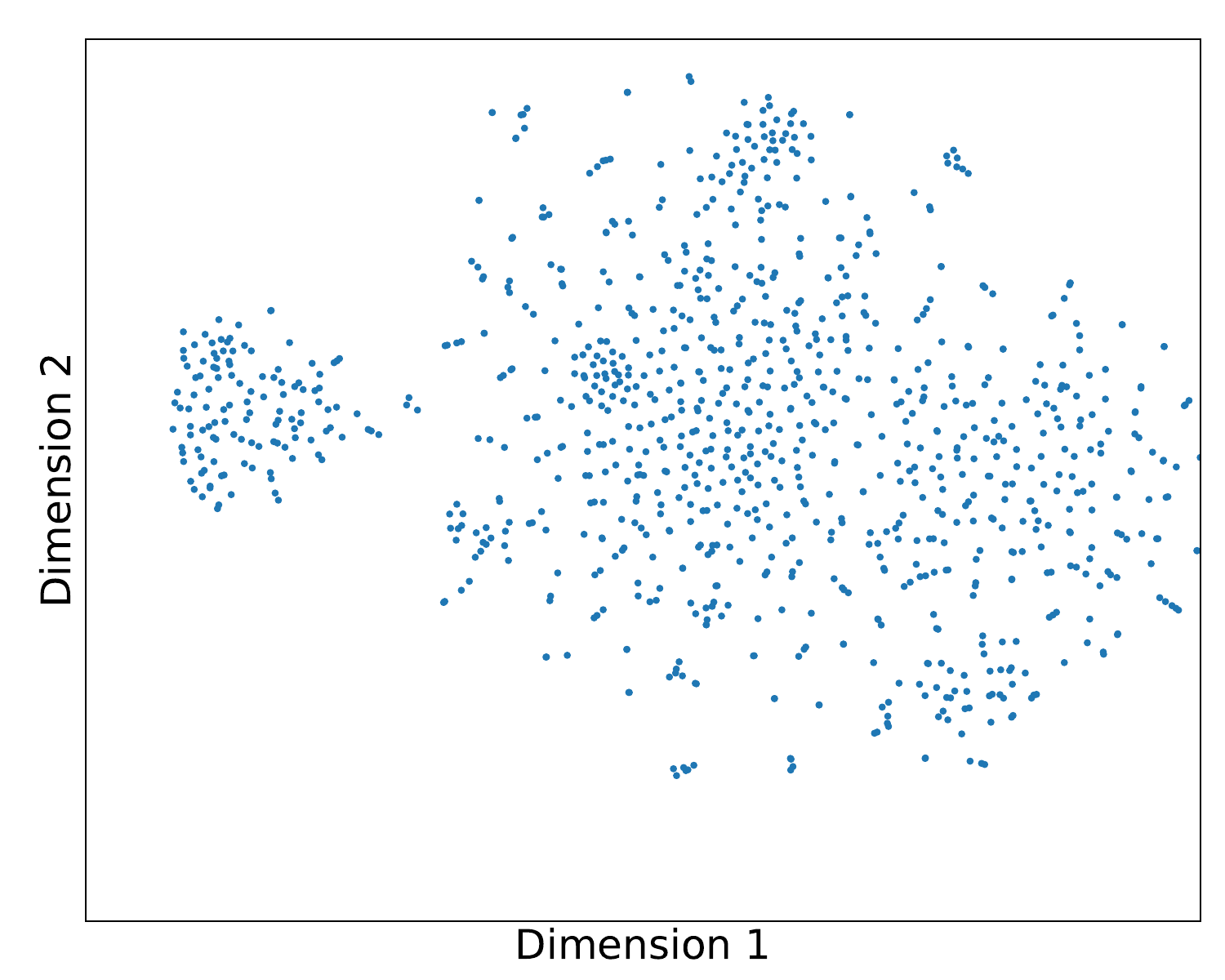}
        \caption{Retraining}
        \label{fig:retraining}
    \end{subfigure}
    \begin{subfigure}[b]{0.19\textwidth}
        \includegraphics[width=\textwidth]{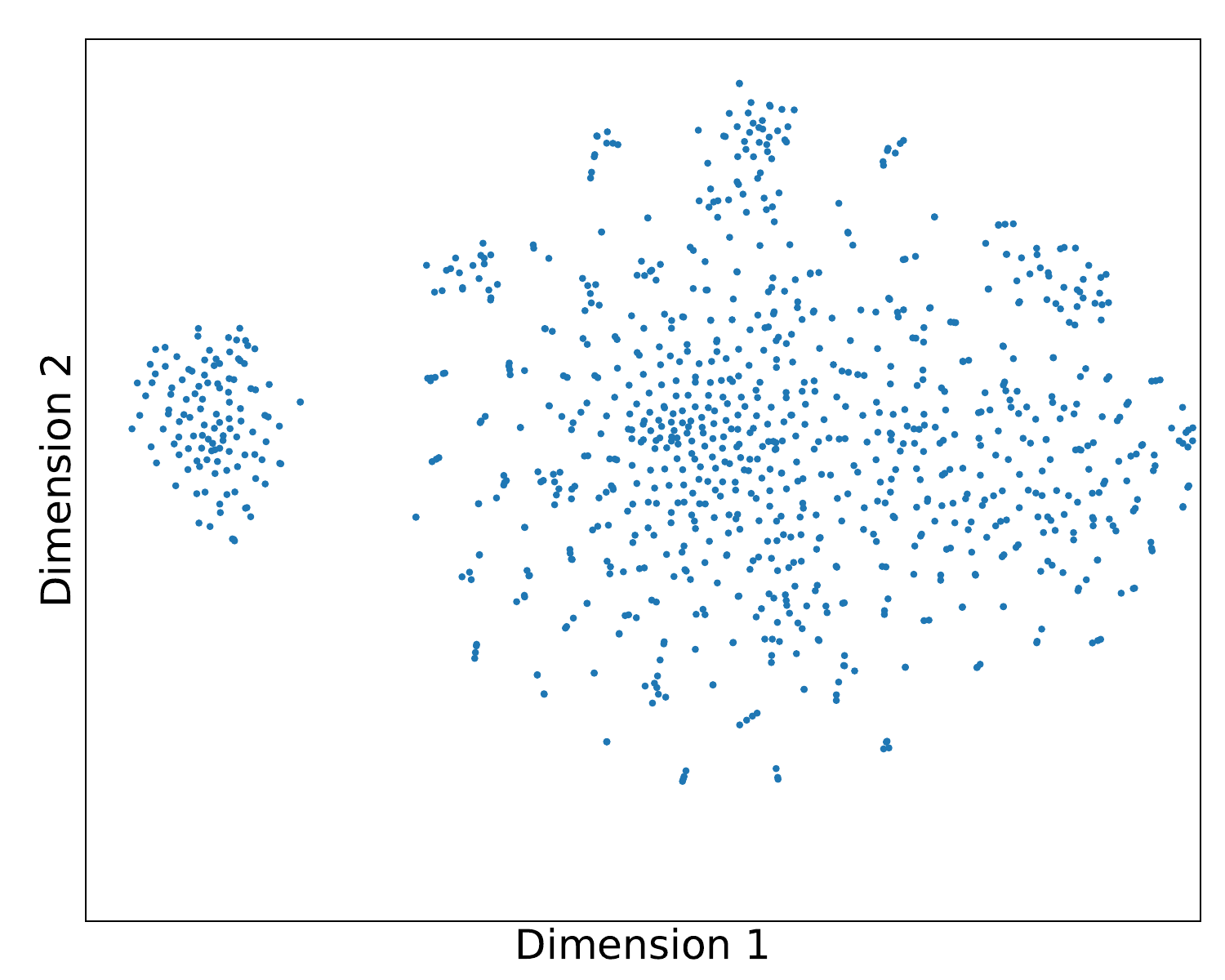}
        \caption{GIF, $l_2$: 10.57}
        \label{GIF}
    \end{subfigure}
    \begin{subfigure}[b]{0.19\textwidth}
        \includegraphics[width=\textwidth]{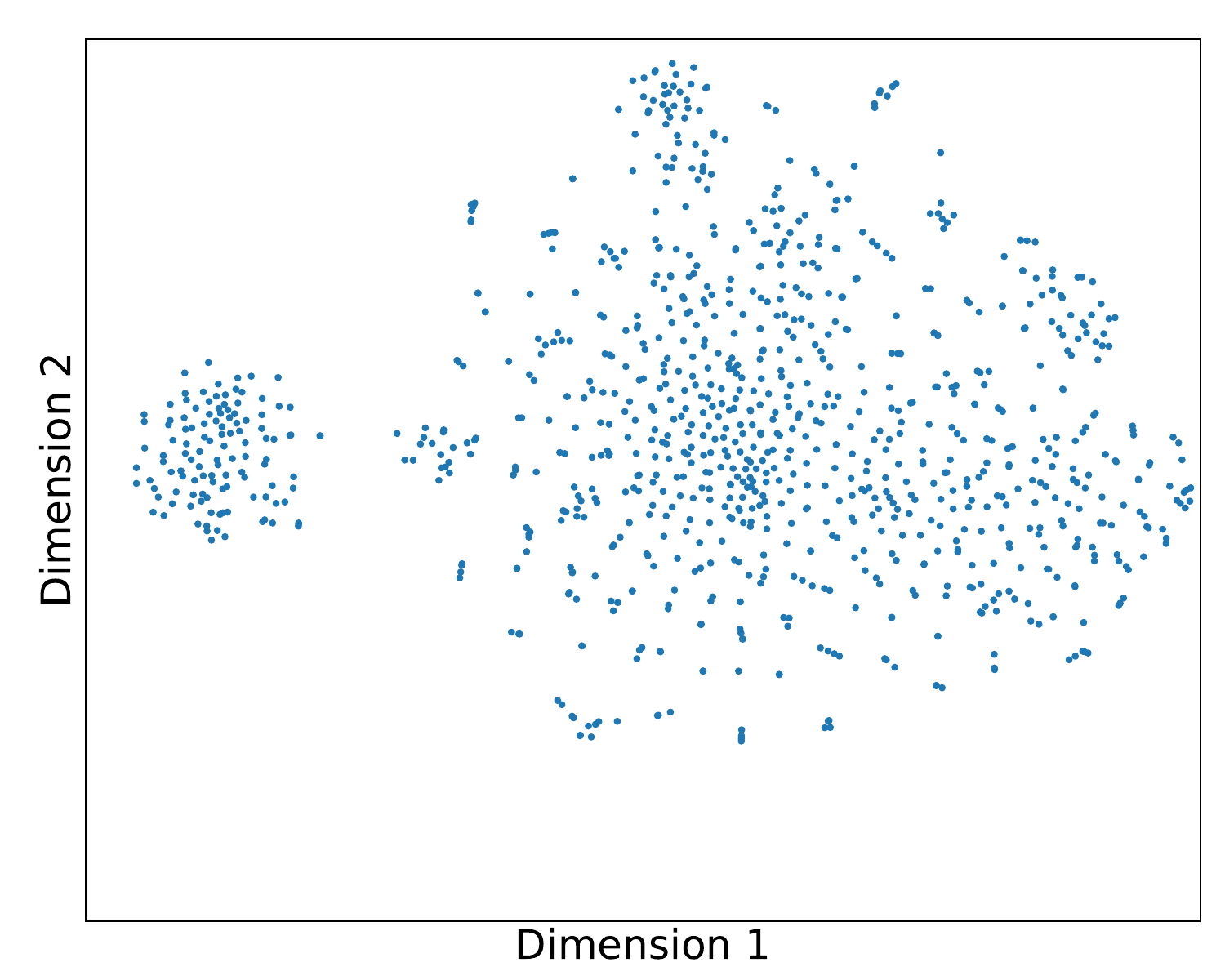}
        \caption{WF-KGIF, $l_2$: 10.56}
        \label{WFGIF}
    \end{subfigure}
    \begin{subfigure}[b]{0.19\textwidth}
        \includegraphics[width=\textwidth]{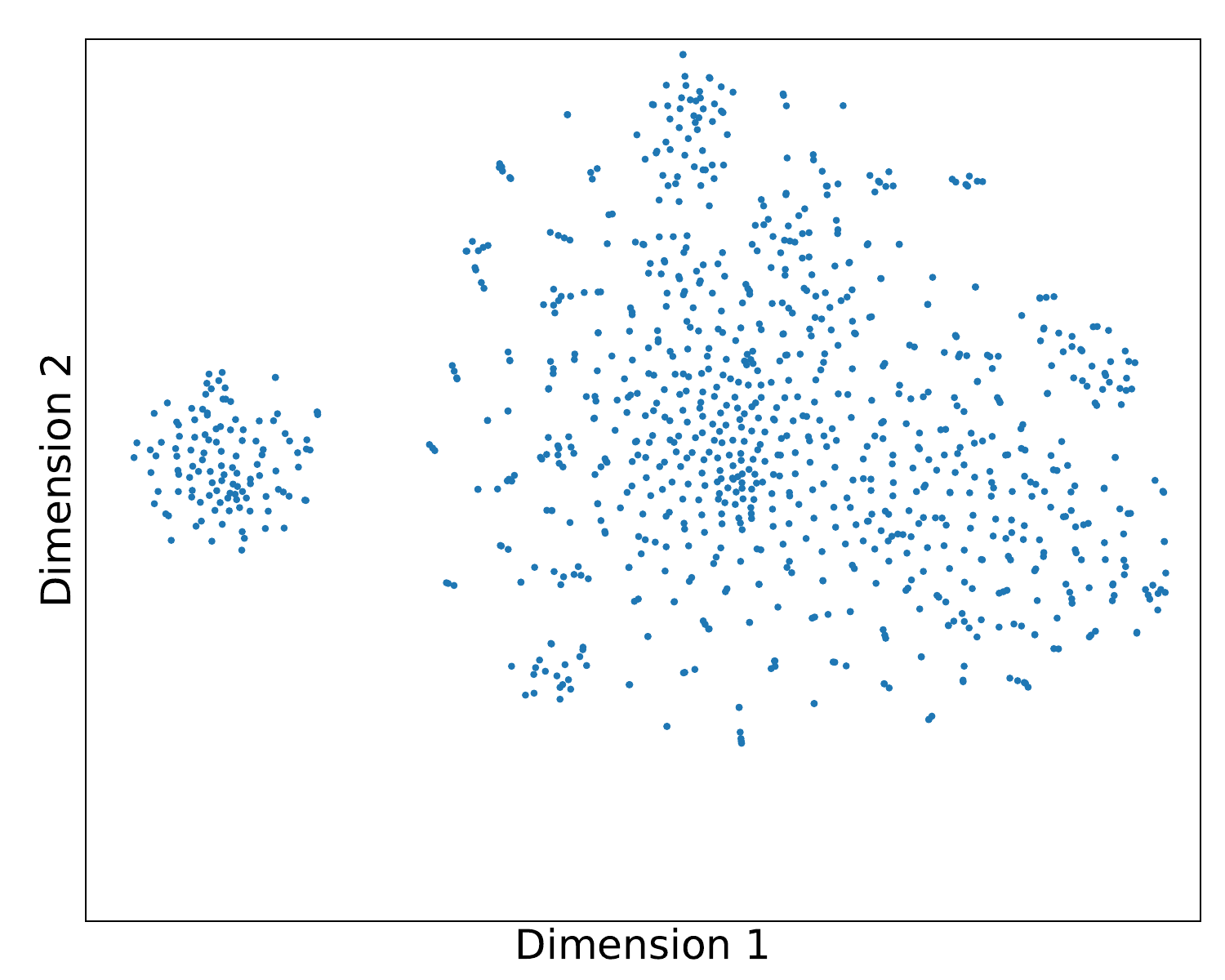}
        \caption{ZeroFisher, $l_2$: 10.53}
        \label{ZOWFGIF}
    \end{subfigure}
    \vspace{-3mm}
    \caption{Visualization of embeddings and L2 distance to retraining}
    \vspace{-3mm}
    \label{fig:visualization}
\end{figure*}

\begin{figure*}[h]
\centering
\begin{subfigure}[b]{0.32\textwidth}
        \includegraphics[width=\textwidth]{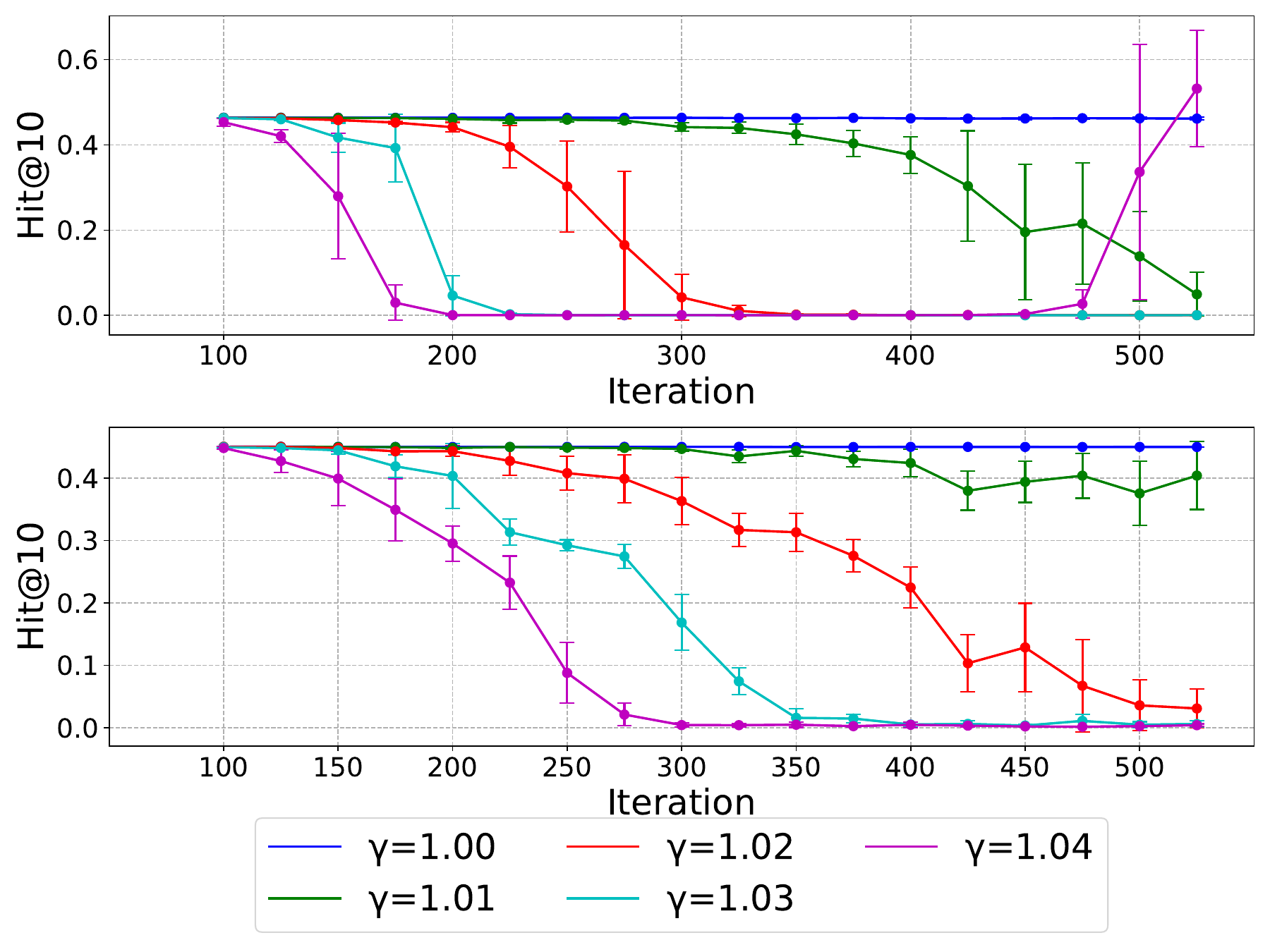}
        \caption{Varying $\gamma$}
        \label{fig:Retain_Models_Iteration-Damp}
    \end{subfigure}
    \begin{subfigure}[b]{0.32\textwidth}
        \includegraphics[width=\textwidth]{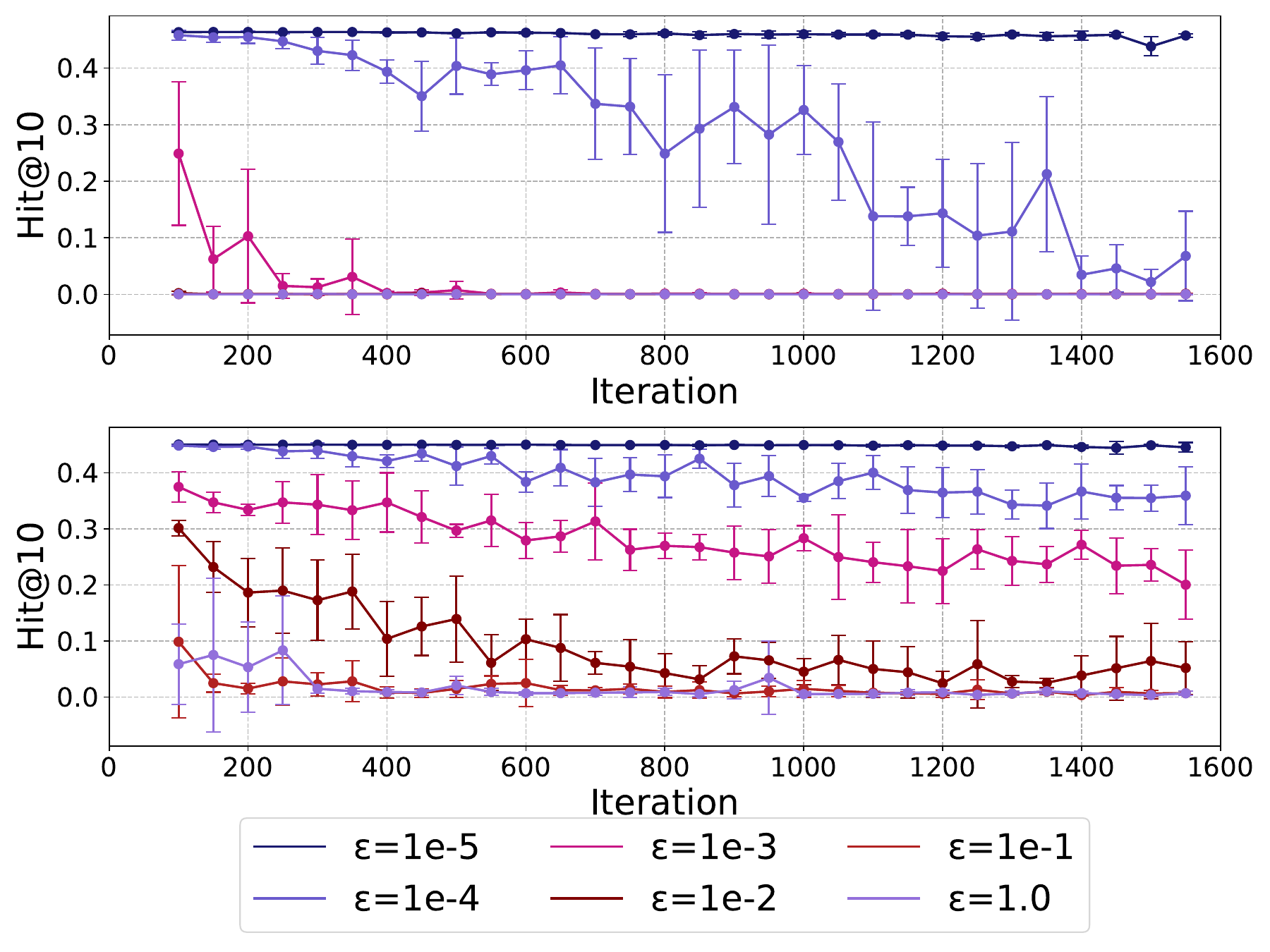}
        \caption{Varying $\epsilon$}
        \label{fig:Retain_Models_Iteration-Epsilon}
    \end{subfigure}
    \begin{subfigure}[b]{0.32\textwidth}
        \includegraphics[width=\textwidth]{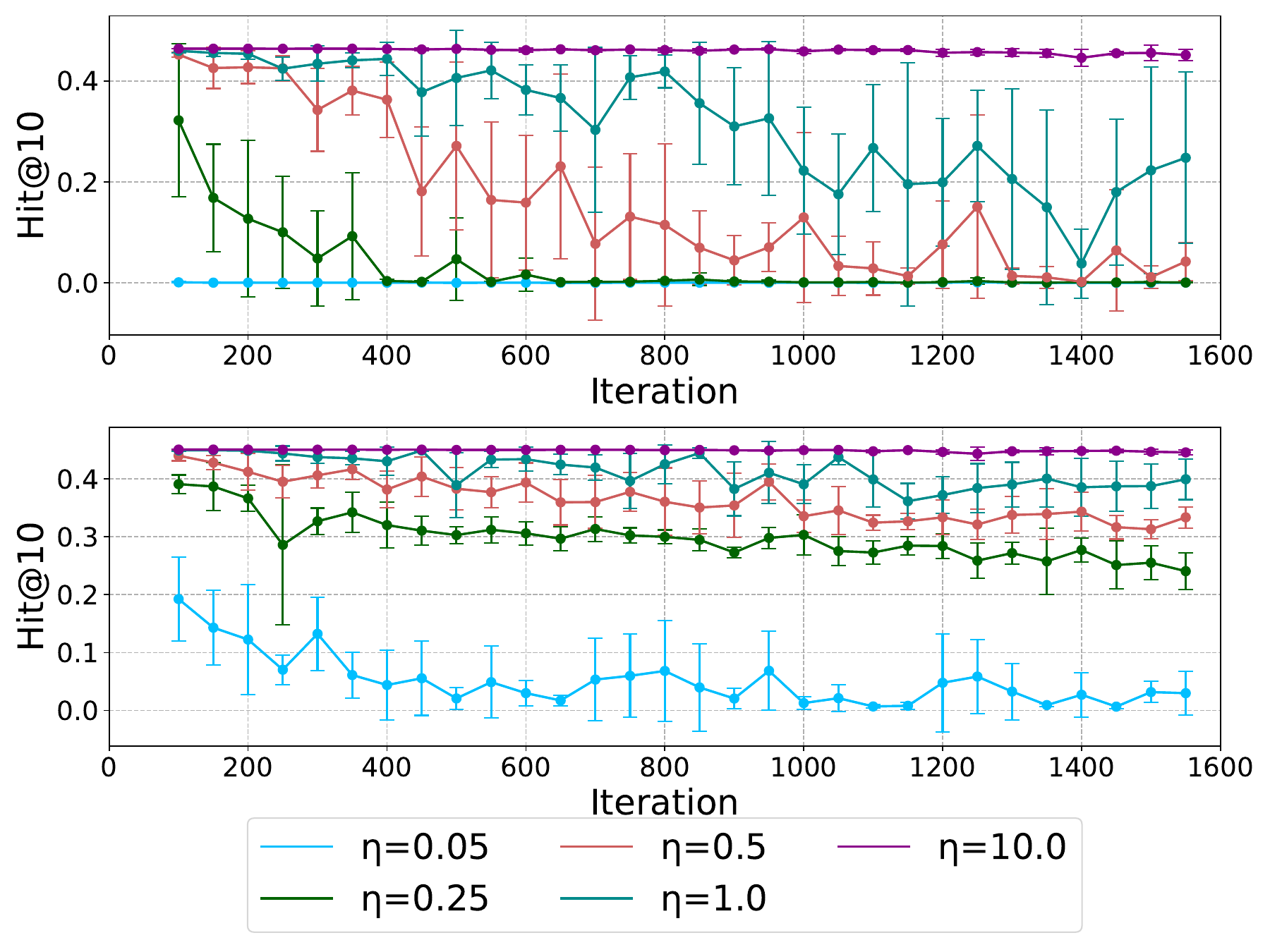}
        \caption{Varying $\eta$}
        \label{fig:Retain_Models_Iteration-Scale}
    \end{subfigure}
    \vspace{-3mm}
    \caption{Varying parameters (Above is RotatE, Below is TransD)}
    \vspace{-3mm}
    \label{fig:Retain_visualization}
\end{figure*}
\begin{table*}[h]
    \centering
    \resizebox{0.9\linewidth}{!}{
    \begin{tabular}{|l|l|cc|ccc|cc|ccc|}
    \hline
    \multirow{2}{*}{Model} &\multirow{2}{*}{Method}&\multicolumn{5}{c|}{5\%}&\multicolumn{5}{c|}{7\%}\\
   & &Time&Memory & MRR & Hit@3 & Hit@1 & Time & Memory& MRR & Hit@3 & Hit@1\\
    \hline
    \multirow{7}{*}{RotatE} & Train & 1666s & 932MB & 0.3165  & 0.3506 & 0.2251 & 1666s & 932MB & 0.3165  & 0.3506 & 0.2251 \\
    \cline{2-12}
    \rowcolor{lightred} 
    \cellcolor{white} & Retrain & 1572s & 901MB & 0.2838 & 0.3191 & 0.1867 & 1596s & 889MB & 0.2751 & 0.3093 & 0.1783 \\
    \cline{2-12}
    & GNNDelete & 247s & 969MB & 0.2746 & 0.3056 & \textbf{0.1841} & 365s & 946MB & 0.2713 & 0.3006 & 0.1813 \\
    \cline{2-12}
    & CertifiedGU & 184s & 9834MB & \textbf{0.2813} & \textbf{0.3162} & 0.1882 & 186s & 9743MB & \textbf{0.2749} & 0.3078 & 0.1826 \\
    \cline{2-12}
    & GIF & 5.82s & 6566MB & 0.2766 & 0.3088 & 0.1799 & 4.66s & 6473MB & 0.2799 & 0.3088 & 0.1799 \\
    \cline{2-12}
    & WF-KGIF (Our) & 1.01s & 2101MB & 0.2779 & 0.3103 & 0.1795 & 1.00s & 2101MB & 0.2706 & 0.3075 & \textbf{0.1795} \\
    \cline{2-12}
    \rowcolor{lightblue}
    \cellcolor{white} & ZeroFisher (Our) & \textbf{0.94s} & \textbf{798MB} & 0.2783 & 0.3113 & 0.1811 & \textbf{0.95s} & \textbf{794MB} & 0.2748 & \textbf{0.3090} & 0.1811 \\
    \hline
    \multirow{7}{*}{TransD} & Train & 1702s & 893MB & 0.2928  & 0.3314 & 0.1971 & 1702s & 893MB & 0.2928  & 0.3314 & 0.1971 \\
    \cline{2-12}
    \rowcolor{lightred}
    \cellcolor{white} & Retrain & 1603s & 888MB & 0.2788 & 0.3156 & 0.1824 & 1586s & 876MB & 0.2745 & 0.3076 & 0.1804 \\
    \cline{2-12}
    & GNNDelete & 379s & 961MB & 0.2737 & 0.3073 & 0.1805 & 371s & 833MB & 0.2700 & 0.3012 & 0.1777 \\
    \cline{2-12}
    & CertifiedGU & 610s & 31769MB & 0.2674 & 0.3053 & 0.1685 & 612s & 31760MB & 0.2615 & 0.2978 & 0.1633 \\
    \cline{2-12}
    & GIF & 7.16s & 21770MB & 0.2729 & 0.3055 & 0.1742 & 9.0s & 21765MB & 0.2569 & 0.2934 & 0.1587 \\
    \cline{2-12}
    & WF-KGIF (Our) & 1.17s & 4540MB & 0.2738 & 0.3068 & 0.1750 & 1.06s & 4540MB & 0.2693 & 0.2944 & 0.1599 \\
    \cline{2-12}
    \rowcolor{lightblue}
    \cellcolor{white} & ZeroFisher (Our) & 1.02s & 768MB & 0.2750 & 0.3068 & 0.1771 & 0.76s & 670MB & 0.2713 & 0.2976 & 0.1631 \\
    \hline
    \multirow{7}{*}{TransH} & Train & 1804s & 773MB & 0.2951 & 0.3315 & 0.2008 & 1804s & 773MB & 0.2951 & 0.3315 & 0.2008 \\
    \cline{2-12}
    \rowcolor{lightred}
    \cellcolor{white} & Retrain & 1650s & 767MB & 0.2787 & 0.3138 & 0.1824 & 1648s & 742MB & 0.2740 & 0.3085 & 0.1648 \\
    \cline{2-12}
    & GNNDelete & 370s & 811MB & 0.2742 & 0.3085 & 0.1790 & 374s & 789MB & 0.2683 & 0.3015 & 0.1742 \\
    \cline{2-12}
    & CertifiedGU & 609s & 30791MB & 0.2696 & 0.2998 & 0.1699 & 580s & 30603MB & 0.2627 & 0.2962 & 0.1663 \\
    \cline{2-12}
    & GIF & 6.74s & 20938MB & 0.2736 & 0.3080 & 0.1764 & 7.03s & 20936MB & 0.2679 & 0.3018 & 0.1683 \\
    \cline{2-12}
    & WF-KGIF (Our) & 1.01s & 4168MB & 0.2737 & 0.3077 & 0.1771 & 1.02s & 4168MB & 0.2688 & 0.3031 & 0.1690 \\
    \cline{2-12}
    \rowcolor{lightblue} 
    \cellcolor{white} & ZeroFisher (Our) & 0.96s & 537MB & 0.2744 & 0.3104 & 0.1800 & 0.98s & 538MB & 0.2697 & 0.3066 & 0.1740 \\    
    
    \hline
    \end{tabular}}
    \caption{Deleting different percentages of triplets on FB15K237.}
    \label{tab:percents_unlearning_performance}
\end{table*}

\begin{figure}[h]
    \centering
    \begin{subfigure}[b]{0.19\textwidth}
    \hspace{-0.275\textwidth}
        \includegraphics[width=1.25\textwidth]{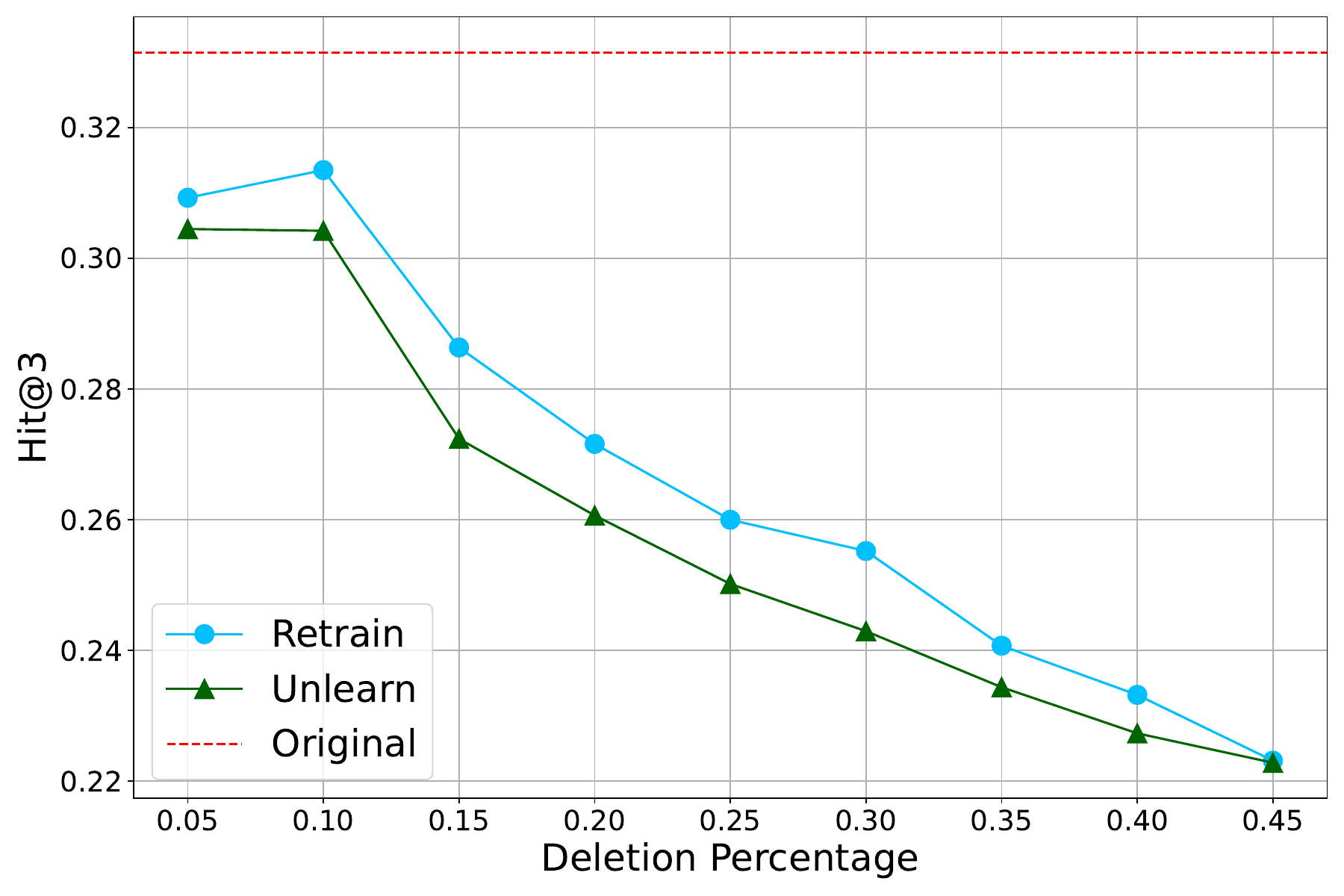}
        \caption{Test Results}
        \label{fig:Deletion_Percentage_Test}
    \end{subfigure}
    \begin{subfigure}[b]{0.19\textwidth}
        \includegraphics[width=1.25\textwidth]{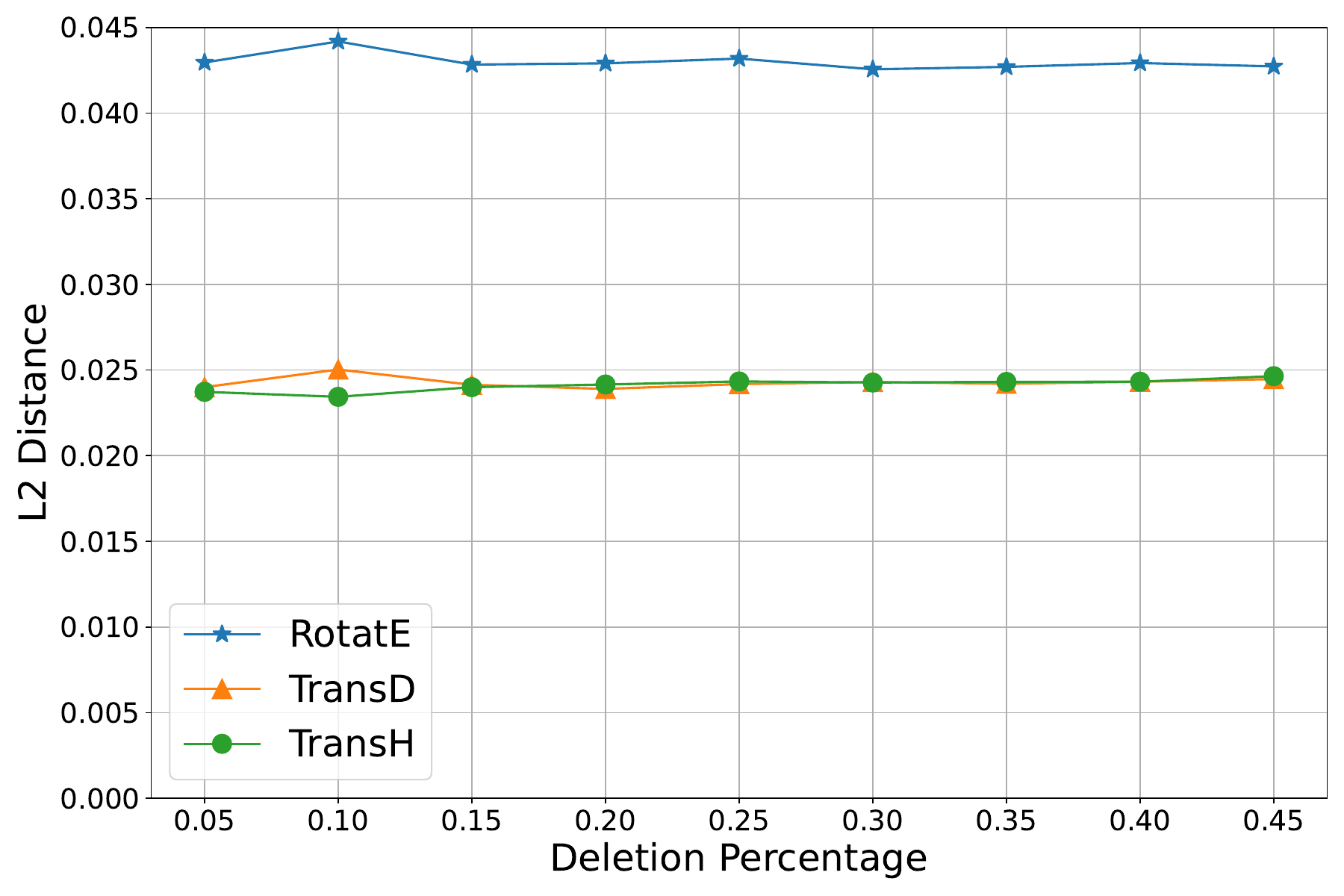}
        \caption{L2 Distance}
        \label{fig:Deletion_Percentage_L2Distance}
    \end{subfigure}
    \vspace{-3mm}
    \caption{Different deletion percentages}
    \vspace{-3mm}
    \label{fig:Deletion_Percentage}
\end{figure}

% Retraining is usually used as the gold standard for evaluating the machine unlearning algorithm~\cite{DBLP:journals/corr/abs-2209-02299}.
% To measure the forget quality, we report the performance of the KG models. With optimal unlearning, the KG performance after unlearning should be close to retraining, as desired. In this paper, we report the performance of the most important downstream task: knowledge graph completion. This task is to predict potential yet unidentified edges, thereby improving the quality of the KG database. We use four evaluation metrics: Mean Reciprocal Rank (MRR), Hit@10, Hit@3, and Hit@1~\cite{DBLP:conf/acl/WangGWK23}.
Retraining is usually used as the gold standard for evaluating the machine unlearning algorithm~\cite{DBLP:journals/corr/abs-2209-02299}.
To measure the forget quality, we report the performance of the KG models. With optimal unlearning, the KG performance after unlearning should be close to retraining, as desired. In this paper, we report the performance of the most important downstream task: knowledge graph completion. This task is to predict potential yet unidentified edges (relationships not explicitly labeled) in the knowledge graph, thereby improving the quality and reliability of the KG database. We use four evaluation metrics: Mean Reciprocal Rank (MRR), Hit@10, Hit@3, and Hit@1. MRR calculates the reciprocal rank of the correct answer in the prediction results for each triplet in the test set and then averages these reciprocal ranks $MRR=\frac{1}{N}\Sigma_{i=1}^{N}\frac{1}{rank_i}$. Hit@k evaluates the percentage at which the top-k prediction results contain the correct triplet in the test set: $Hit@k = \frac{1}{N} \Sigma_{i=1}^{N} \mathbf{1} (rank_i \leq k)$, where $\mathbf{1}$ refers indicator function.

\subsection{Results and Analysis.} 
\subsubsection{Results.}
Talbe~\ref{table: results} presents the comparison between our algorithm and baselines in terms of efficiency and quality for triplet unlearning.
Our unlearning KG algorithm based on approximation with the Fisher matrices and Zeroth-order information (ZeroFisher) significantly outperforms the baselines in terms of runtime and memory usage. We also include one variant of our algorithms which computes the gradients directly instead of approximating with zeroth-order information.
Specifically, the training time of ZeroFisher is the smallest among all methods. For example, the runtime of ZeroFisher for RotatE is 0.48s while the runtime of the best baseline (GIF) is 9.3s. Also, our memory usage is only about 25\% of GIF across all KG models and datasets. This is because our method is the only one that does not require constructing the computational graph to compute the derivatives. While retraining is the gold standard to measure the unlearning quality, our method archives the closest performance to the retraining result in terms of most metrics. 
Table~\ref{tab:entities_unlearning_performance} and~\ref{tab:edges_unlearning_performance} are results of entity and relation unlearning. It shows similar conclusion. The ZeroFisher method is memory-efficient, with significantly lower memory usage compared to other methods. Although some metrics in some method dataset combinations were slightly lower than the optimal method, the performance of the ZeroFisher method was still very close to the optimal value. This indicates that our method can provide good performance while maintaining high efficiency. A reasonable explanation is that our approximation leverages the empirical Fisher matrices which are calculated based on the distribution of the remaining data. The overall results indicate that our method is efficient while guaranteeing the unlearning quality.

\subsubsection{Ablation Study.}
% \noindent\textbf{Ablation Study}

To verify the effectiveness of each proposed approximation method, we conduct an ablation study. Specifically, we introduce two variants of our method: WF-KGIF and KGIF. Instead of using Zeroth-order information to estimate gradients, WF-KGIF only leverages the Woodbury theorem and the Fisher matrices for unlearning.  KGIF further directly computer the IHVPs. In Figure~\ref{fig:ablation}, we show the runtime and the peak value of the GPU memory usage of three variants on two datasets. The performance of WF-KGIF is also included in Table~\ref{table: results}. Note that we have recorded the memory usage in the three important steps of computing the influence function: loss computing, gradient computing, and IHVPs. We can see that approximation with the Woodbury theorem and the Fisher matrices can reduce the runtime significantly. Zoreth-order information can further reduce the runtime. We also observe that IHVPs account for the majority of memory usage. For IHVPs, gradient computation, and loss value computation, replacing our approximation with direct computation will increase memory usage. It verifies the necessity of each proposed component.

\vspace{-3mm}
\subsubsection{Effects of Hyperparameters.}
% \noindent\textbf{Effects of Hyperparameters.}
We investigate the impact of 4 hyperparameters on the unlearning results: iterations, the sampling term, the scaling down term, and the noise term. In Figure~\ref{fig:Varying}, we visualize Hit@10 for different configurations. More results are included in~\textbf{Appendix B}.

\begin{table*}[!htp]
    \centering
    \resizebox{0.95\linewidth}{!}{
    \begin{tabular}{|l|l|cc|ccc|cc|ccc|}
    \hline
    \multirow{2}{*}{Model} &\multirow{2}{*}{Method}&\multicolumn{5}{c|}{FB15K237}&\multicolumn{5}{c|}{YAGO3}\\
   & &Time&Memory & MRR & Hit@3 & Hit@1 & Time & Memory& MRR & Hit@3 & Hit@1\\
    \hline
    \multirow{7}{*}{RotatE} & Train & 1666s & 932MB & 0.3165  & 0.3506 & 0.2251 & 11049s & 4493MB& 0.3381 & 0.3846 & 0.2356 \\
    \cline{2-12}
    \rowcolor{lightred} 
    \cellcolor{white} & Retrain & 1342s & 784MB & 0.2913 & 0.3217 & 0.2100 & 9672s & 3278MB & 0.3249 & 0.3707 & 0.2261 \\
    \cline{2-12}
    & GNNDelete & 337s & 856MB & 0.2867 & 0.3159 & 0.2043 & 1572s & 3677MB & 0.2799 & 0.3278 & 0.2003 \\
    \cline{2-12}
    & CertifiedGU & 182s & 9504MB & 0.2944 & 0.3227 & 0.2070 & 201s & 39768MB & 0.3216 & 0.3667 & 0.2245 \\
    \cline{2-12}
    & GIF & 3.5s & 6399MB & 0.2935 & 0.3211 & 0.2066 & 11.9s & 26417MB & 0.3202 & 0.3652 & 0.2265 \\
    \cline{2-12}
    & WF-KGIF (Our)  & 0.23s & 2040MB & 0.2954 & 0.3142 & 0.1997 & 0.50s & 9032MB & 0.3217 & 0.3679 & 0.2209 \\
    \cline{2-12}
    \rowcolor{lightblue} 
    \cellcolor{white} & ZeroFisher (Our) & 0.19s & 774MB & 0.2978 & 0.3175 & 0.2079 & 0.48s & 4125MB & 0.3304 & 0.3738 & 0.2290 \\
    \hline
    \multirow{7}{*}{TransD} & Train & 1702s & 893MB & 0.2928  & 0.3314 & 0.1971 & 8543s & 3642MB & 0.3197  & 0.4542 & 0.2796\\
    \cline{2-12}
    \rowcolor{lightred} 
    \cellcolor{white} & Retrain & 1478s & 796MB & 0.2696 & 0.3056 & 0.1828 & 7907s & 3471MB & 0.3442 & 0.4003 & 0.2381 \\
    \cline{2-12}
    & GNNDelete & 455s & 850MB & 0.2600 & 0.2928 & 0.1733 & 1499s & 3517MB & 0.2894 & 0.3467 & 0.2015 \\
    \cline{2-12}
    & CertifiedGU & 529s & 31459MB & 0.2833 & 0.3177 & 0.1910 & - & OOM & - & - & - \\
    \cline{2-12}
    & GIF & 6.6s & 21221MB & 0.2643 & 0.2984 & 0.1805 & 18.6s & 84079MB & 0.3398 & 0.4071 & 0.2265 \\
    \cline{2-12}
    & WF-KGIF (Our)  & 0.29s & 4066MB & 0.2603 & 0.2908 & 0.1794 & 1.97s & 19718MB & 0.3325 & 0.3986 & 0.2156 \\
    \cline{2-12}
    \rowcolor{lightblue} 
    \cellcolor{white} & ZeroFisher (Our)& 0.36s & 628MB & 0.2681 & 0.2924 & 0.1862 & 1.86s & 3235MB & 0.3482 & 0.4076 & 0.2372 \\    
    \hline
    \multirow{7}{*}{TransH} & Train & 1804s & 773MB & 0.2951  & 0.3315 & 0.2008 & 8421s & 3118MB & 0.4212  & 0.4851 & 0.3108\\
    \cline{2-12}
    \rowcolor{lightred} 
    \cellcolor{white} &  Retrain & 1435s & 688MB & 0.2697 & 0.3043 & 0.1840 & 7851s & 2969MB & 0.3522 & 0.4104 & 0.2443 \\
    \cline{2-12}
    & GNNDelete & 439s & 730MB & 0.2736 & 0.3057 & 0.1865 & 1513s & 3143MB & 0.2837 & 0.3334 & 0.1917 \\
    \cline{2-12}
    & CertifiedGU & 489s & 29453MB & 0.2850 & 0.3176 & 0.1942 & - & OOM & - & - & - \\
    \cline{2-12}
    & GIF & 6.3s & 20409MB & 0.2658 & 0.2992 & 0.1721 & 16.7s & 82465MB & 0.3548 & 0.4133 & 0.2499\\
    \cline{2-12}
    & WF-KGIF (Our)  & 0.38s & 4038MB & 0.2641 & 0.2935 & 0.1736 & 1.62s & 19718MB & 0.3576 & 0.4146 & 0.2508 \\
    \cline{2-12}
    \rowcolor{lightblue} 
    \cellcolor{white} & ZeroFisher (Our)& 0.32s & 487MB & 0.2708 & 0.2951 & 0.1879 & 1.37s & 2614MB & 0.3494 & 0.4023 & 0.2434 \\    
    
    \hline
    \end{tabular}
    }
    \vspace{2mm}
    \caption{Relation Unlearning on FB15K237 and YAGO3.}
    \vspace{-4mm}
    \label{tab:edges_unlearning_performance}
\end{table*}

Each iteration updates the parameter with estimation, bringing them closer to the desired state. A higher number of iterations typically leads to convergent results, as the algorithm has more opportunities to refine the parameter estimates. Recall that the damping term in Eq.\ref{Woodfisher_hvps} plays a vital role in controlling the update magnitude, ensuring that the update process is neither too aggressive nor too slow, thereby balancing convergence speed and stability. As the damping term increases (see Figure~\ref{fig:TransH_Iteration-Damp}),  the performance gradually becomes unstable, and the standard deviation increases. This is because if the damping term is too high, it will lead to an excessive increase in the estimation of IHVPs. As the number of iterations increases, the overall parameter estimation will accumulate, resulting in less robust results.

The scaling-down term determines how much the calculated updates are scaled down before being applied to the model parameters. A lower value of the scaling down term results in greater fluctuations in the parameter updates, as the adjustments are more sensitive to each iteration's computation. Conversely, a higher value for the scaling down term results in smaller, more controlled updates, making the training process more stable. The noise term $\epsilon$ is critical for determining the accuracy of the approximating gradients. When the noise term is set to a smaller value, the perturbations are smaller, leading to more precise gradient approximations. Consequently, the performance of the unlearning process improves, resulting in model parameters that are closer to the desired state. A larger noise term speeds up the convergence but at the cost of performance drop.

\begin{table*}[!htp]
    \centering
    \small
    \begin{tabular}{|c|c|c|c|c|c|c|c|c|c|}
    \hline
         Method& Type& Dataset& Del. Ratio& Unlearn MRR& Unlearn Hit@3& Unlearn Hit@1& Remain MRR& Remain Hit@3& Remain Hit@1\\ \hline
         Original& Random& FB15K237& 10\%& 0.5166& 0.6067& 0.3798& 0.4708& 0.5536& 0.3339
\\
         Retrain& Random& FB15K237& 10\%& 0.3247& 0.3856& 0.2008& 0.5094& 0.5920& 0.3768
\\
         ZeroFisher& Random& FB15K237& 10\%& 0.4875& 0.5030& 0.3101& 0.4707& 0.5536& 0.3339
\\\hline
         Original& Random& YAGO3-10& 10\%& 0.5662& 0.6337& 0.4503& 0.5588& 0.6270& 0.4420
\\
         Retrain& Random& YAGO3-10& 10\%& 0.2800& 0.3210& 0.1792& 0.5769& 0.6425& 0.4625
\\
         ZeroFisher& Random& YAGO3-10& 10\%& 0.5056& 0.5379& 0.3879& 0.5581& 0.6259& 0.4411\\
    \hline
    \end{tabular}
    \caption{Discussion about the data distribution}
    \label{tab:distribution}
\end{table*}

\subsubsection{Embeddings Distance.}
To further investigate the unlearning quality, we visualize the embedding difference after unlearning. We randomly sample 1,000 entities and project their embeddings into 2-dimensional space with PAC. By comparing original embeddings (see Figure \ref{fig:original}) and embeddings after unlearning (see Figure \ref{fig:retraining} to~\ref{ZOWFGIF}), we found that the removal of data can have an impact on the distribution of embeddings. We also show the $L_2$ distance between embedding after retraining and the ones after approximate unlearning. Compared with the best baseline (GIF), the embedding after unlearning with our algorithm is closer to retraining. It explicitly proves the quality of our unlearning method.

% \vspace{-6mm}
\subsubsection{Deletion Percentage.}
In Figure~\ref{fig:Deletion_Percentage}, we investigate the impact of deletion percentage.  As the deletion percentage increases, The performance of the KG model drops too(See in Figure~\ref{fig:Deletion_Percentage_Test}). However, our unlearning method consistently remains close to the retraining. Also, $L_2$ The $L2$ distance between embedding after retraining and the ones with our unlearning remains stable as the deletion percentage increases. More experiments are dicussed in~\ref{sec:appendix:A}.

% \vspace{-8mm}
\section{Discussion}
All datasets used in this paper are publicly available. Our research can be easily reproduced. Unfortunately, there is currently no strong evaluation metrics in Machine Unlearning that indicates whether the model weights have completely removed the influence of some data. For the retraining model, the deleted dataset actually represents a special validation set or test set, because retraining model have never seen the deleted dataset just like the validation set or test set. The worst performance on the deleted dataset does not mean that it is truly forgotten. In contrast, because of the generalization of the retraining model itself, it can also achieve good results on the deleted set, even if the deleted set has never participated in training process. MUSE~\cite{shi2024muse} use Member Inference Attack (MIA)~\cite{kim2024detecting} to classify if a data is in training set. But~\cite{duan2024membership} found that MIA barely outperform random guessing for most settings across varying LLM sizes and domains. Back to the question itself, how to determine whether the model has unlearned this data?~\cite{zheng2025spurious} proposed that what we truly unlearn is how to use the knowledge but the knowledge itself. They showed how deleted data affects the distribution of the model from the loss landscape. So this might be an out-of-distribution problem. Ignoring the default test set, we select original, retrain and our method models to test on the remaining data set and the deleted set. The unlearned model deviates from the original model and moves towards the retrained model (See in tab~\ref{tab:distribution}). Introducing distribution alignment or optimal transport~\cite{xiao2025optimal} into machine unlearning could yield broader conclusions across KGs, CV, NLP, and other data modalities.

% \vspace{-6mm}
\section{Conclusion}
This paper introduces a novel influence function–based framework to effectively eliminate the residual effects of deleted triplets, entities, and relations in knowledge graph (KG) models. Recognizing the high computational cost of traditional influence function computation, we propose a hierarchical approximation strategy to improve efficiency without sacrificing unlearning performance. Specifically, we first approximate the second-order inverse Hessian–vector products (IHVPs) using tractable first-order information, significantly reducing the need for costly Hessian computations. To further improve scalability, we approximate the first-order gradients using zeroth-order information through carefully designed perturbation-based estimations. This two-step approximation makes the influence estimation process both scalable and model-agnostic. We validate the effectiveness and efficiency of our method through extensive experiments on standard benchmark datasets, comparing against both full retraining and existing unlearning baselines. Results demonstrate that our approach achieves competitive unlearning quality while significantly reducing computational overhead, offering a practical solution for safe and efficient knowledge removal in deployed KG systems.

\appendix

\vspace{-3mm}
\section{Appendix A}
\label{sec:appendix:A}

Table~\ref{tab:percents_unlearning_performance} is the result of deleting different percentages of triplets. As the deletion percentage increases, the performance of the KG model drops. However, our unlearning method consistently remains close to the retraining. Also, deleting more data will increase the running time of unlearning for all unlearning methods. Compared with the baselines, our unlearning method is especially efficient while deleting more data. 

\definecolor{lightblue}{RGB}{217,233,243}
\definecolor{lightgray}{gray}{0.9}

\vspace{-3mm}
\section{Appendix B}
\label{sec:appendix:B}
Figrue~\ref{fig:Retain_visualization} shows hyper-parameter combination results on two models. 
The damping term controls update magnitude, balancing convergence speed and stability.
As the damping term increases, performance becomes unstable due to inflated IHVP estimates.
As the number of iterations increases, the overall parameter estimation will accumulate, resulting in less robust results. 
% The scaling-down term determines how much the calculated updates are scaled down before being applied to the model parameters. 
The scaling-down term controls how much updates are reduced before applying to model parameters.
% A lower value of the scaling down term results in greater fluctuations in the parameter updates, as the adjustments are more sensitive to each iteration's computation. 
A lower scaling-down value causes larger parameter fluctuations, making updates more sensitive to each iteration.
A higher value for the scaling down term results in smaller, more controlled updates, making the training process more stable. 
The noise term is critical for determining the accuracy of the approximating gradients. 
When the noise term is set to a smaller value, the perturbations are smaller, leading to more precise gradient approximations. 
Consequently, the performance of the unlearning process improves, resulting in model parameters that are closer to the desired state. A larger noise term speeds up the convergence, but at the cost of a performance drop.

\section{GenAI Usage Disclosure}
This document was fully created and edited by humans. AI tools have not been used to create the content.

% This paper introduces an influence function to eliminate the effects of deleted triplets, entities, and relations on KG models. Recognizing the high computational cost of traditional influence function computation, we propose a hierarchical approximation strategy to improve efficiency without sacrificing unlearning performance. Specifically, we first approximate the second-order IHVPs with the first-order information. Then we further estimate the first-order information with zeroth-order information. The efficiency and unlearning quality of our method are verified by comparing retraining and baselines on Benchmark datasets. 
\balance
\bibliographystyle{ACM-Reference-Format}
\bibliography{acmart}

\end{document}